\documentclass[10pt,twocolumn,letterpaper]{article}

\usepackage[pagenumbers]{eojepa}  

\usepackage{multirow}
\usepackage{colortbl}

\usepackage{microtype}

\makeatletter
\renewcommand\paragraph{\@startsection{paragraph}{4}{\z@}{4pt plus 1pt minus 1pt}{-1em}{\normalfont\normalsize\bfseries}}
\makeatother

\providecommand{\tbd}{\textcolor{red}{\textbf{TBD}}}

\definecolor{linkblue}{rgb}{0.21,0.49,0.74}
\usepackage[pagebackref,breaklinks,colorlinks,allcolors=linkblue]{hyperref}
\usepackage{tabularx}
\usepackage{array}

\title{EO-JEPA: Learning Long-Horizon Dynamics of Earth Observation}

\author{%
Diaa Addeen Abuhani\textsuperscript{1,2} \quad
Mohamed Atef Abouelhadid\textsuperscript{1} \quad
Muhammad Akhtar Munir\textsuperscript{1} \quad
Hiyam Debary\textsuperscript{3} \\
Fahad Shahbaz Khan\textsuperscript{1,4} \quad
Salman Khan\textsuperscript{1,5,*} \\[6pt]
{\normalsize \textsuperscript{1}Mohamed bin Zayed University of Artificial Intelligence \quad \textsuperscript{2}Massachusetts Institute of Technology} \\
{\normalsize \textsuperscript{3}IBM \quad \textsuperscript{4}Link\"{o}ping University \quad \textsuperscript{5}Australian National University} \\[4pt]
{\small \textsuperscript{*}Corresponding author} \\[4pt]
{\small Code, weights and project page: \url{https://mbzuai-oryx.github.io/EO-JEPA}}
}

\begin{document}
\maketitle
\begin{abstract}
Forecasting land-surface responses to weather over months to years remains challenging, particularly during droughts and heat extremes. Pixel-space models devote capacity to unpredictable visual variations, such as clouds and atmospheric noise, and often drift towards seasonal patterns during long autoregressive rollouts. We introduce EO-JEPA, which formulates earth observation forecasting as action-conditioned latent world modeling, representing the land surface as a latent state and meteorology as an exogenous conditioning signal. Predictions evolve entirely in latent space, with pixels decoded only at the output. EO-JEPA achieves the highest ENS among learned models on Extreme Summer and the 200- and 700-day EarthNet2021 tracks, improving by 1.7, 0.8, and 4.4 points, respectively. It also doubles the correlation between predicted and observed inter-annual vegetation changes under contrasting weather. Our work establishes latent-state rollout as a practical alternative to pixel-space autoregression for forecasting weather-dependent land-surface dynamics over extended horizons.
\end{abstract}    
\section{Introduction}
\label{sec:intro}

\begin{figure}
    \centering
    \includegraphics[width=\linewidth]{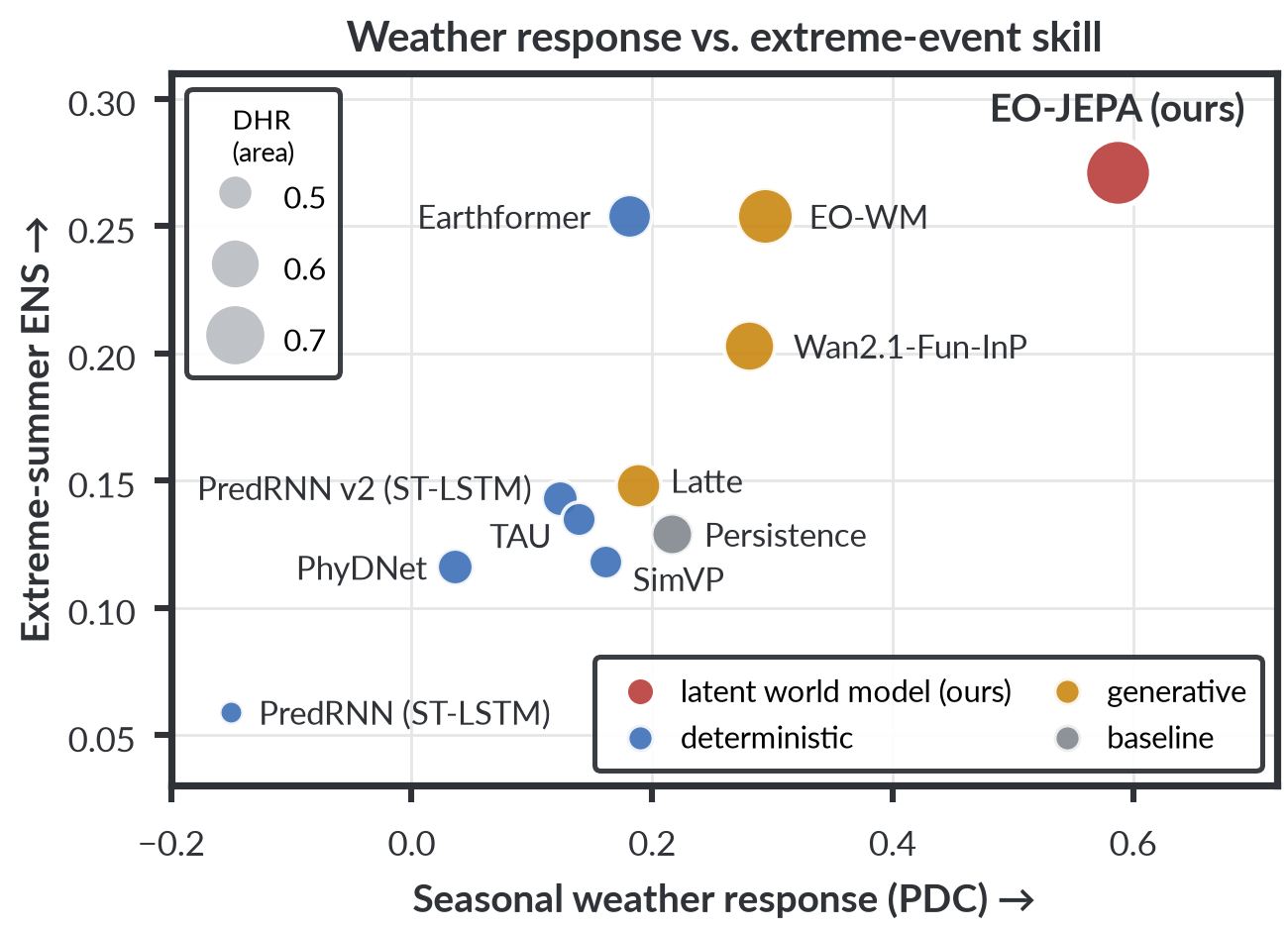}
    \caption{EO-JEPA achieves twice the pair-difference correlation (PDC) of prior models on the Seasonal benchmark and the highest EarthNetScore (ENS) on Extreme Summer, demonstrating strong performance in both seasonal weather response and extreme-event forecasting. Bubble area represents direction hit rate (DHR).}
    \label{fig:hook}
\end{figure}

Satellite observations provide repeated, high-resolution coverage of the Earth's surface, capturing environmental and climate-driven changes over time \cite{drusch2012sentinel2, reichstein2013extremes,bastos2020legacy}. Recent advances in spatiotemporal deep learning have enabled forecasting future observations from historical imagery at the pixel level \cite{requena2021earthnet,gao2022earthformer,benson2024multimodal}, often incorporating auxiliary future meteorological signals \cite{diaconu2022understanding,robin2022learning,benson2024multimodal}. However, forecasting remains challenging over extended horizons, particularly under extreme conditions, when accurate predictions are arguably more critical \cite{ciais2005europe,requena2021earthnet,kladny2024enhanced}.

As the forecast horizon grows, predictions tend to drift towards the typical seasonal cycle \cite{requena2021earthnet}, limiting their ability to capture the magnitude of vegetation departures from it during droughts \cite{kladny2024enhanced,pellicervalero2025explainable,luo2026eowm}. We argue that this stems partly from forecasting in pixel space. Satellite imagery contains substantial variability from clouds, atmospheric noise, and fine-scale textures \cite{wilson2016cloud,skakun2022cmix}, much of which is inherently unpredictable over long horizons. 

Pixel-level objectives allocate model capacity to transient visual details rather than to the underlying dynamics, contributing to error accumulation during autoregressive rollouts and a drift towards typical seasonal behavior. We hypothesize that the predictable component of land-surface dynamics, particularly vegetation responses to weather, can be represented in a lower-dimensional space than the pixels used to observe it. This motivates forecasting in a compact latent space that prioritizes these underlying dynamics over pixel-level reconstruction.

Inspired by advances in latent representation learning, we explore forecasting in a compact space that preserves predictable structure while discarding noisy visual detail. Joint-Embedding Predictive Architectures (JEPA) provide one way to learn such representations \cite{lecun2022path, assran2023ijepa, assran2025vjepa2}. When conditioned on actions, they further enable world models with stable, controllable rollouts over long horizons \cite{maes2026leworldmodel}.

We formulate Earth-surface forecasting as action-conditioned latent world modeling, representing the land surface as a latent state and future meteorology as the exogenous input driving its evolution. First, satellite images contain unpredictable details, including clouds, shadows, illumination and sensor noise. Pixel reconstruction expends capacity on these details and feeds errors into subsequent autoregressive steps. We therefore roll out forecasts in the latent spaces of frozen pretrained encoders, whose tokens de-emphasize these details, and decode images only at the output. Second, a latent space suited to modeling dynamics may not be optimal for image synthesis. Semantic embeddings describe land cover and vegetation but cannot be decoded, whereas autoencoder latents support decoding but must retain appearance. We therefore predict complementary semantic and decodable streams to represent dynamics and generate images. Third, land surfaces respond to weather with delayed and cumulative effects, and during extremes, forecast weather may differ from that in the context frames. We condition every step on weather as an explicit action, making its relationship with land-surface responses testable. As in the EarthNet protocol, predictions use observed ERA5-Land weather during the forecast period, measuring land-surface prediction under known weather rather than operational forecasting. Our contributions are threefold:

\begin{itemize}
  \item We introduce EO-JEPA, a weather-conditioned latent world model that combines frozen semantic and decodable representations to predict land-surface dynamics, with autoregressive rollout performed entirely in latent space and image decoding deferred to the output.
  \item We demonstrate strong performance under observed future weather on extreme-event and long-horizon benchmarks, improving EarthNetScore (ENS) over the strongest reported learned baselines by 0.8 and 4.4 points on the 200- and 700-day EarthNet2021 tracks, respectively.
  \item We evaluate weather-response fidelity alongside image-level accuracy. EO-JEPA approximately doubles the correlation of predicted and observed interannual vegetation differences relative to EO-WM \cite{luo2026eowm}, while weather-input controls show that conventional accuracy metrics alone can obscure substantial losses in response fidelity.
\end{itemize}
On the 700-day track, EO-JEPA closes most of the 4.9-point gap between the best published learned model and persistence, and it lowers the trough-NDVI error of the generative baselines in every drought-severity tier. This positions latent world modeling as a strategy for forecasting the Earth's surface over long horizons and under extremes.

\section{Related Work}
\label{sec:related_works}
\begin{figure*}[t]
    \centering
\includegraphics[width=0.9\linewidth]{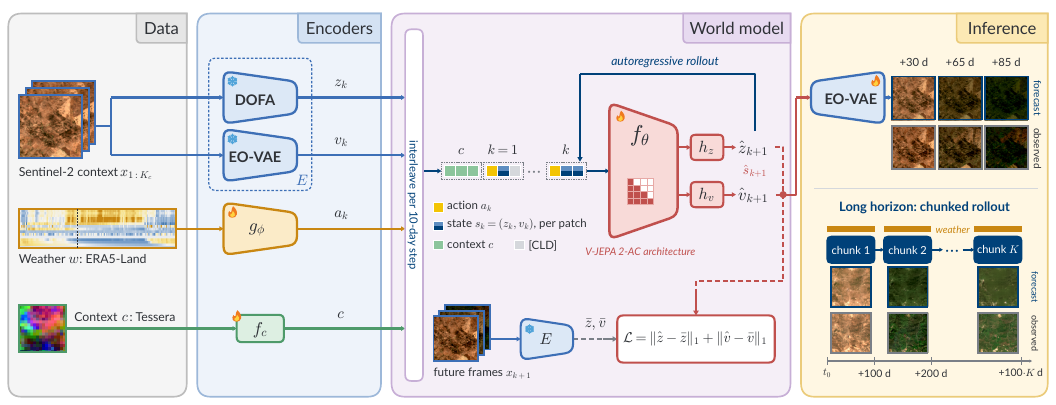}
\caption{\textbf{EO-JEPA overview.}
Frozen DOFA and EO-VAE encoders extract complementary semantic ($z_k$) and decodable ($v_k$) tokens from Sentinel-2 observations. ERA5-Land weather is encoded into actions $a_k$, while prior-year Tessera embeddings provide context $c$. A block-causal predictor $f_\theta$ models weather-conditioned latent dynamics over interleaved context, action, and state tokens, predicting future states through teacher-forced and rollout training. Only predicted decodable tokens $\hat v_k$ are decoded into images, preventing pixel-space errors from propagating. Long-horizon forecasts use chunked rollout, initializing each 100-day chunk from the previous chunk's final $K_c$ predicted states. The bottom row shows predictions and observations at the ends of chunks 1, 2, and $K$ over the 700-day seasonal track.}

    \label{fig:architecture}
\end{figure*}
\paragraph{Earth Observation Forecasting}


Satellite image forecasting is commonly framed as guided video prediction, using historical observations alongside static topography and, in some cases, dynamic meteorological variables \cite{requena2021earthnet,robin2022learning,benson2024multimodal}. Existing approaches predict primarily in pixel space using convolutional recurrent networks \cite{shi2015convlstm,diaconu2022understanding,robin2022learning,kladny2024enhanced}, spatiotemporal transformers \cite{gao2022earthformer,benson2024multimodal}, video-prediction architectures \cite{guen2020phydnet,gao2022simvp,wang2022predrnn,tan2023tau}, and generative video models \cite{ma2024latte,wan2025,luo2026eowm}. Despite promising results at conventional horizons, performance often degrades over multiple seasons \cite{requena2021earthnet,diaconu2022understanding,kladny2024enhanced}.

We instead forecast in the latent spaces of frozen pretrained encoders, decode only at the output, and treat weather as an action. Closely relevant, EO-WM~\cite{luo2026eowm} uses weather-conditioned latent diffusion, while Sat-JEPA-Diff~\cite{komurcu2026sat} predicts semantic features but feeds generated images back into its rollout. EO-JEPA instead predicts and feeds back complementary semantic and decodable latent states at every step, avoiding pixel-space feedback while enabling long-horizon evaluation of vegetation responses to weather.

\paragraph{Latent Predictive Representations}

JEPA-based models learn representations by predicting the embeddings of masked or future content rather than reconstructing it at the pixel level, allowing them to focus on predictable structure without modeling inherently unpredictable details \cite{lecun2022path,assran2023ijepa,bardes2024vjepa,assran2025vjepa2}. In earth observation, this is particularly relevant as many variations in satellite imagery arise from transient factors, such as clouds and atmospheric conditions, rather than meaningful changes in the underlying land surface.

Self-supervised and multi-sensor foundation models produce general-purpose representations through a range of learning objectives, including masked image modeling \cite{cong2022satmae,jakubik2023prithvi,fuller2023croma,tseng2025galileo}, cross-modal representation learning \cite{wang2024decur}, wavelength-conditioned encoding \cite{xiong2024dofa}, JEPA-style predictive learning \cite{lundqvist2025geojepa,astruc2025anysat,choudhury2026xjepa}, generative multimodal modeling \cite{jakubik2025terramind}, and pixel-level temporal embedding learning \cite{feng2025tessera}.

These models learn representations of observed imagery for downstream tasks; even JEPA-style approaches predict embeddings of masked content within an observation rather than how the state evolves under external influences. We use pretrained encoders as a frozen state space and learn weather-conditioned dynamics within it. 



\paragraph{Action-Conditioned World Models}
World models learn how an environment evolves given an agent's actions, and are used for control and planning, in image space \cite{ha2018world,hafner2019planet,hafner2023dreamerv3,bruce2024genie,hu2023gaia} or on top of frozen pretrained features \cite{zhou2024dinowm,assran2025vjepa2,bar2025navigation}.
The actions there are low-dimensional, controllable and take effect immediately, and rollouts span seconds to minutes.

We adopt the block-causal, action-conditioned latent predictor used in these models, but adapt it to a different type of input and temporal relationship. Rather than an agent command, our action represents weather forcing, an exogenous and uncontrollable driver that most strongly affects forecast performance. Since weather effects are delayed and cumulative, a causal weather encoder summarizes the 90 days preceding the forecast window in the first action, and the predictor attends to the full action history. We also extend rollouts to hundreds of days, using action conditioning not for planning but to make land-surface responses to weather explicit and testable through counterfactual evaluations.

\definecolor{figEnc}{HTML}{3D6FB6}   
\definecolor{figMet}{HTML}{C98712}   
\definecolor{figCtx}{HTML}{4E9A6A}   
\definecolor{figPred}{HTML}{C0504D}  
\newcommand{\Cenc}[1]{\textcolor{figEnc}{#1}}
\newcommand{\Cmet}[1]{\textcolor{figMet}{#1}}
\newcommand{\Cctx}[1]{\textcolor{figCtx}{#1}}
\newcommand{\Cpred}[1]{\textcolor{figPred}{#1}}

\section{Methodology}
\label{sec:3_methodology}

We consider a spatiotemporal cube of satellite observations $x_{1:T}$, with $x_t \in \mathbb{R}^{H \times W \times B}$ a Sentinel-2 image with $B$ spectral bands, acquired every five days over a fixed location and accompanied by a per-pixel cloud mask. We group consecutive frames into steps of ten days, indexed by $k$, so that a forecasting window consists of $K_c$ observed context steps followed by $K_f$ steps to be predicted. Alongside the imagery, each location provides daily meteorology $\Cmet{w_{1:T'}}$ and static or slowly varying context $\Cctx{c}$, given by a land-surface embedding from the preceding year.

As seen in Fig.~\ref{fig:architecture}, we cast forecasting as \emph{action-conditioned} latent world modeling. Rather than predicting future pixels, we represent the land surface at step $k$ by a latent state
\begin{equation}
  \Cenc{s_k} = \Cenc{E}(x_{k}),
\end{equation}
where $\Cenc{E}$ denotes two frozen pretrained encoders applied to the frames of step $k$, giving a semantic and a decodable stream, $\Cenc{s_k} = (\Cenc{z_k}, \Cenc{v_k})$ (details in Sec.~\ref{sec:latent}). The evolution of the state is driven by an action $\Cmet{a_k}$, the meteorological forcing that governs the transition from step $k$ to step $k{+}1$:
\begin{equation}
\begin{gathered}
  \Cmet{h_d} = \Cmet{g_\phi}\big(\Cmet{w_{d-12:d}}\big), \quad
  \Cmet{a_k} = W_a\big[\operatorname{mean}_{\mathcal{D}_k} \Cmet{h} ;\, \max_{\mathcal{D}_k} \Cmet{h}\big],\\
  \mathcal{D}_1 = (\tau_0 - L,\, \tau_2], \quad \mathcal{D}_k = (\tau_k,\, \tau_{k+1}] \;\; (k > 1),
\end{gathered}
\label{eq:action}
\end{equation}
where $\Cmet{h_d}$ is the feature of day $d$, computed by the learned encoder $\Cmet{g_\phi}$ (Sec.~\ref{sec:action}) from that day's weather and, through causal temporal convolutions, the 12 preceding days, and $W_a$ is a layer-normalized linear map of the element-wise mean and maximum over the days $\mathcal{D}_k$. $\tau_k$ is the last day of step $k$ and $\tau_0$ the day before the first step of the current 150-day window, so the first action also pools the $L{=}90$ antecedent days. Later actions pool only their own transition; earlier weather reaches them through the predictor, which attends to all previous actions. Unlike actions in robotics or games, $\Cmet{a_k}$ is exogenous and uncontrollable, and its effect on the surface unfolds over weeks rather than instantaneously. A predictor $\Cpred{f_\theta}$ models the transition dynamics in latent space,
\begin{equation}
  \Cenc{\hat{s}_{k+1}} = \Cpred{f_\theta}\big(\Cenc{s_{\le k}},\, \Cmet{a_{\le k}},\, \Cctx{c}\big),
\end{equation}
attending only to past states and actions. Forecasts beyond the observed context are obtained by autoregressive rollout, feeding each prediction back as input for the next step:
\begin{equation}
  \Cenc{\hat{s}_{k+1}} = \Cpred{f_\theta}\big(\Cenc{s_{1:K_c}},\, \Cenc{\hat{s}_{K_c+1:k}},\, \Cmet{a_{\le k}},\, \Cctx{c}\big), \qquad k \ge K_c .
\end{equation}
Only at the output is the predicted state mapped back to reflectance by a decoder, $\hat{x}_{k} = \Cenc{D}(\Cenc{\hat{s}_{k}})$. Throughout the rollout, no pixels are reconstructed, so model capacity is devoted to the evolution of the land surface rather than to transient visual detail such as clouds, atmospheric effects and fine-scale texture. We test this choice directly by routing the rollout through pixel space, decoding and re-encoding every predicted step with the same model (Sec.~\ref{sec:experiments} and Sec.~\ref{supp:rollout}).

This formulation separates three questions that pixel-space forecasting entangles; (a) what the surface state is (fixed by $\Cenc{E}$), (b) how it evolves under weather (learned by $\Cpred{f_\theta}$ and $\Cmet{g_\phi}$), and (c) how it appears in an image (handled by $\Cenc{D}$). It also makes the role of weather explicit and testable as $\Cmet{a_k}$ enters only through the action pathway, its contribution can be isolated by withholding it, or replacing it with the weather of another location, at inference, which we exploit in supplementary Sec.~\ref{supp:weather}.

\subsection{Latent State}
\label{sec:latent}


The state combines two frozen encoders with complementary roles: a semantic stream $\Cenc{z_k}$, which the predictor forecasts but which is never decoded, and a decodable stream $\Cenc{v_k}$, from which the output images are rendered.

\paragraph{Semantic stream.} The semantic stream $\Cenc{z_k}$ is from DOFA \cite{xiong2024dofa}, a ViT-B/16 foundation model whose patch embedding is generated from the wavelength of each input band, so it ingests bands without channel re-mapping or zero-filling. To retain both low-level spectral and structural information, we concatenate the patch tokens of blocks 4, 8 and 12, giving $\Cenc{z_k^p} \in \mathbb{R}^{2304}$ per patch $p$. A step contains two acquisitions; we combine their features per token, weighting each frame by the fraction of cloud-free pixels in the patch, so that a clear observation is not diluted by a cloudy one. Features are layer-normalized per token. Although $\Cenc{z_k}$ is never decoded, it carries the representation in which the dynamics are learned.

\paragraph{Decodable stream.} The decodable stream, $\Cenc{v_k}$, consists of the latent tokens of EO-VAE \cite{eovae}, a variational autoencoder for multispectral imagery whose decoder renders the forecast (Sec.~\ref{sec:decoding}). We adapt EO-VAE once to the training imagery with a reconstruction and NDVI-consistency objective, as the released weights distort the red/near-infrared ratio on which vegetation indices depend, and keep it frozen. Unlike $\Cenc{z_k}$, the two acquisitions of a step are kept separate, $\Cenc{v_k^p} \in \mathbb{R}^{2 \times 128}$, so the forecast retains the native five-day revisit.

\paragraph{Clouds.} A token is observed in acquisition $j$ ($o_{k,j}^p{=}1$) when at least half of its pixels there are cloud-free, and in step $k$ ($m_k^p{=}1$) when at least half of its pixels in both acquisitions together are. Unobserved tokens are never imputed. Semantic features of unobserved steps are replaced with a learned \texttt{[CLD]} embedding, and EO-VAE tokens of unobserved acquisitions are zeroed and excluded from the loss, so a patch clear in only one acquisition takes both streams from that acquisition.



\subsection{Weather as Action}
\label{sec:action}
Vegetation responds to weather with delays of weeks to months, integrating forcing over time \cite{vicente2013response}. The action encoder $\Cmet{g_\phi}$ captures these accumulated and lagged effects. Inputs are 15 daily ERA5-Land \cite{munozsabater2021era5land} variables on a $40 \times 40$ grid, covering weather and 4 soil-moisture layers (supplementary Tab.~\ref{tab:supp_weather}). Standardized using training-set statistics, the inputs span 90 antecedent days and the forecasting window.

The encoder extracts spatial weather features for every day with convolutional layers, followed by causal temporal convolutions that give each day's feature a memory of the preceding 12 days. Each action pools these features over its days $\mathcal{D}_k$ (Eq.~\ref{eq:action}). In chunked rollouts (Sec.~\ref{sec:longhorizon}), each chunk is a standard window with its own actions from the observed weather; its $\tau_0$ is the day before its first context step, so its first action pools the 90 observed days before the chunk.

\subsection{Context}
\label{sec:context}

The context $\Cctx{c}$ supplies slow and static information that the observed steps may not reveal, particularly under clouds.

\paragraph{Land-surface prior.} We use Tessera \cite{feng2025tessera} embeddings, which summarize a full year of Sentinel-1/2 observations per 10\,m pixel into a 128-dimensional vector. We take the embedding of the year preceding the cube, so it contains no information from the forecast period, average it to the $16{\times}16$ token grid, and project it to $P$ prefix tokens. Patches without valid embeddings receive a learned null token. The prefix tells the model what kind of surface each patch is (e.g. cropland) and how it behaved over a seasonal cycle, information that ten 5-day observations cannot provide. We deliberately provide no geographic coordinates, so the model cannot memorize region-specific behavior and must infer dynamics from land cover and the observed state.

\paragraph{Elevation and date.} A digital elevation model of the cube, embedded by an $8{\times}8$ patch convolution, is added to every state token of the patch, and each step carries a date token, an MLP of the day of year (as $\sin$/$\cos$) and the year. Both have little effect on the forecasts (supplementary Tab.~\ref{tab:supp_static}).

\paragraph{Condition dropout.} During training, the action sequence, the Tessera prefix, the elevation embedding and the date token are each replaced by learned null embeddings with probability 0.15, independently per sample. This makes every input individually removable at inference, which we use to measure what each contributes (supplementary Sec.~\ref{supp:weather}).

\subsection{Block-Causal Predictor}
\label{sec:predictor}
The predictor $\Cpred{f_\theta}$ (25M parameters) follows the action-conditioned architecture of V-JEPA~2-AC~\cite{assran2025vjepa2}, using a transformer over interleaved action and state tokens (Fig.~\ref{fig:architecture}):

\begin{equation}
  \big[\, \Cctx{c} \;\big|\; \{\Cmet{a_k},\, e_k^{1:P}\}_{k=1}^{K} \,\big],
\end{equation}

where $\Cctx{c}$ denotes the $P$ Tessera prefix tokens, and each patch state is encoded as

\begin{equation}
  e_k^p = W_z\, \Cenc{\tilde{z}_k^p} + m_k^p\, W_v\, \mathrm{LN}\big(\big[o_{k,1}^p \Cenc{v_{k,1}^p} ;\, o_{k,2}^p \Cenc{v_{k,2}^p}\big]\big),
\end{equation}
where $\Cenc{\tilde{z}_k^p}$ equals $\Cenc{z_k^p}$ for observed tokens and $\texttt{[CLD]}$ otherwise, $m_k^p$ and $o_{k,j}^p$ are the step and acquisition masks, and $W_v$ maps the concatenated EO-VAE vectors of the two acquisitions ($2{\times}128$ values) to the model width; predicted steps count as observed. Learned type embeddings distinguish token types, while 3D rotary embeddings encode (step, row, column) positions. Action tokens use only the step index, and prefix tokens are placed at step 0.

Attention is \emph{block-causal} where
every token of step $k$ attends to the prefix and to all tokens of steps $\le k$, but not to the future. The outputs at the state tokens of step $k$ are mapped by two linear heads to the next state,
\begin{equation}
  \Cenc{\hat{z}_{k+1}^p} = \Cpred{h_z}(o_k^p), \qquad \Cenc{\hat{v}_{k+1}^p} = \Cpred{h_v}(o_k^p),
\end{equation}
where $o_k^p$ is the final hidden state. 
\begin{figure*}[t]
\centering
\includegraphics[width=0.95\linewidth]{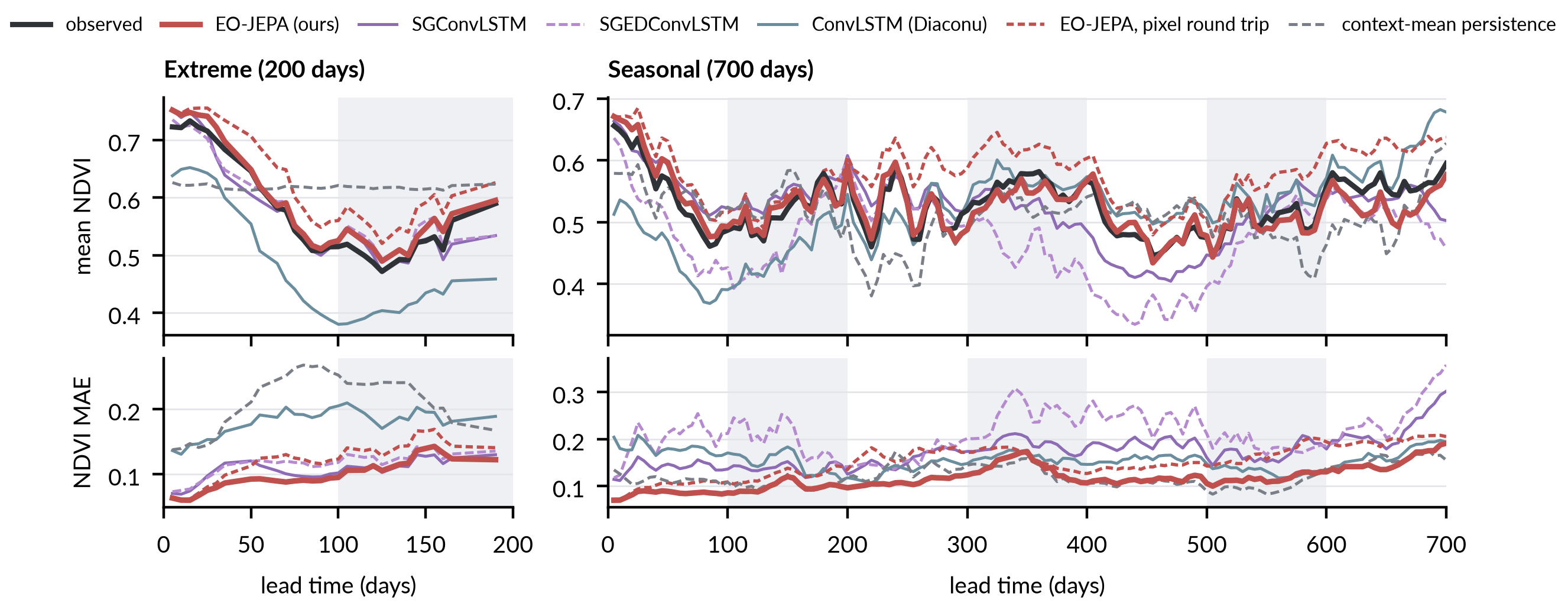}
\caption{Following the observed land surface over 200 and 700 days. Top: mean NDVI per target frame (1{,}000 random cubes per track, cloud-free pixels of the observation; 3-frame moving average); bottom: NDVI MAE. Published models use released weights; the pixel round trip decodes and re-encodes every predicted step of EO-JEPA. Shaded bands mark alternate 100-day chunks; lead times with under 10\% cloud-free pixels are omitted. Supplementary Fig.~\ref{fig:supp_lead} adds the latent ConvLSTM and last-pixel persistence.}
\label{fig:lead}
\end{figure*}

\begin{table*}[t]
\centering
\caption{\textbf{EarthNet2021 long-horizon tracks} (official EarthNetScore). Extreme: 20 context $\rightarrow$ 40 target frames (200 days); Seasonal: 70 $\rightarrow$ 140 (700 days). Published as reported. Earthformer and EO-WM are not run: they report no long-horizon results, their code predicts one 20-frame window, and Earthformer's checkpoint is no longer downloadable. EO-JEPA: raw chunked rollout of 100-day chunks, no blending with persistence; seed 0, over 3 seeds Extreme $0.282 \pm 0.002$, Seasonal $0.262 \pm 0.001$. Bold: best; underline: second best; gray rows are not ranked. Seasonal anchor: per pixel, the nearest cloud-free context frame at the same day of the year ($\pm$15 days), else the context mean.}
\label{tab:long}
\setlength{\tabcolsep}{4pt}
\resizebox{\linewidth}{!}{%
\begin{tabular}{l|ccccc|ccccc}
\toprule
\multirow{2}{*}{Model} & \multicolumn{5}{c|}{Extreme (200 days)} & \multicolumn{5}{c}{Seasonal (700 days)} \\
 & ENS$\uparrow$ & MAD$\uparrow$ & OLS$\uparrow$ & EMD$\uparrow$ & SSIM$\uparrow$ & ENS$\uparrow$ & MAD$\uparrow$ & OLS$\uparrow$ & EMD$\uparrow$ & SSIM$\uparrow$ \\
\midrule
Channel-U-Net \cite{requena2021earthnet}
& \cellcolor{cyan!15}0.236 & \cellcolor{cyan!20}0.229 & \cellcolor{cyan!20}0.297 & \cellcolor{cyan!17}0.206 & \cellcolor{cyan!12}0.231
& \cellcolor{cyan!16}0.196 & \cellcolor{cyan!20}0.217 & \cellcolor{cyan!25}0.381 & \cellcolor{cyan!20}0.190 & \cellcolor{cyan!13}0.126 \\

Arcon \cite{requena2021earthnet}
& \cellcolor{cyan!9}0.222 & \cellcolor{cyan!15}0.224 & \cellcolor{cyan!5}0.275 & \cellcolor{cyan!12}0.198 & \cellcolor{cyan!8}0.208
& \cellcolor{cyan!5}0.159 & \cellcolor{cyan!5}0.201 & \cellcolor{cyan!24}0.379 & \cellcolor{cyan!15}0.179 & \cellcolor{cyan!5}0.083 \\

ConvLSTM \cite{diaconu2022understanding}
& \cellcolor{cyan!5}0.214 & \cellcolor{cyan!5}0.214 & \cellcolor{cyan!16}0.291 & \cellcolor{cyan!5}0.188 & \cellcolor{cyan!5}0.190
& \cellcolor{cyan!22}0.219 & \cellcolor{cyan!18}0.215 & \cellcolor{cyan!23}0.378 & \cellcolor{cyan!24}0.200 & \cellcolor{cyan!20}0.169 \\

SGConvLSTM \cite{kladny2022deep}
& \cellcolor{cyan!31}\underline{0.274} & \cellcolor{cyan!28}\underline{0.237} & \cellcolor{cyan!35}\underline{0.320} & \cellcolor{cyan!32}\underline{0.228} & \cellcolor{cyan!31}\underline{0.350}
& \cellcolor{cyan!21}0.216 & \cellcolor{cyan!23}\underline{0.221} & \cellcolor{cyan!21}0.376 & \cellcolor{cyan!12}0.172 & \cellcolor{cyan!23}\underline{0.182} \\

SGEDConvLSTM \cite{kladny2022deep}
& \cellcolor{cyan!25}0.260 & \cellcolor{cyan!21}0.230 & \cellcolor{cyan!32}0.316 & \cellcolor{cyan!26}0.219 & \cellcolor{cyan!23}0.299
& \cellcolor{cyan!11}0.179 & \cellcolor{cyan!10}0.206 & \cellcolor{cyan!5}0.358 & \cellcolor{cyan!5}0.154 & \cellcolor{cyan!12}0.122 \\

\textbf{EO-JEPA (ours)}
& \cellcolor{cyan!35}\textbf{0.282} & \cellcolor{cyan!35}\textbf{0.244} & \cellcolor{cyan!35}\textbf{0.320} & \cellcolor{cyan!35}\textbf{0.233} & \cellcolor{cyan!35}\textbf{0.374}
& \cellcolor{cyan!35}\textbf{0.263} & \cellcolor{cyan!35}\textbf{0.234} & \cellcolor{cyan!35}\textbf{0.392} & \cellcolor{cyan!35}\textbf{0.228} & \cellcolor{cyan!35}\textbf{0.252} \\
\midrule
\textcolor{gray}{Persistence \cite{requena2021earthnet} (context mean)}
& \textcolor{gray}{0.194} & \textcolor{gray}{0.216} & \textcolor{gray}{0.281} & \textcolor{gray}{0.161} & \textcolor{gray}{0.161}
& \textcolor{gray}{0.268} & \textcolor{gray}{0.233} & \textcolor{gray}{0.385} & \textcolor{gray}{0.203} & \textcolor{gray}{0.318} \\

\textcolor{gray}{Seasonal anchor (training-free, ours)}
& \textcolor{gray}{--} & \textcolor{gray}{--} & \textcolor{gray}{--} & \textcolor{gray}{--} & \textcolor{gray}{--}
& \textcolor{gray}{0.275} & \textcolor{gray}{0.235} & \textcolor{gray}{0.382} & \textcolor{gray}{0.250} & \textcolor{gray}{0.272} \\
\bottomrule
\end{tabular}}
\end{table*}

Since $\Cmet{a_k}$ describes the transition $k \to k{+}1$, each prediction is conditioned on its own weather. All one-step predictions come from a single forward pass; longer forecasts append $\Cenc{\hat{s}_{k+1}}$ as the next, observed, state.


\subsection{Training}
\label{sec:training}

\paragraph{Objective.} Targets are the frozen encoders' outputs for the true next step, $\Cenc{\bar{s}_{k+1}} = (\Cenc{\bar{z}_{k+1}}, \Cenc{\bar{v}_{k+1}})$, eliminating the need for stop-gradient or EMA targets. As in JEPA \cite{bardes2024vjepa,assran2025vjepa2}, the loss is computed in representation space rather than on pixels. We minimize a masked $\ell_1$ loss:
\begin{equation}
  \mathcal{L}_{\text{step}}
  = \sum_{q \in \{\Cenc{z},\Cenc{v}\}} \lambda_q\,
  \frac{\sum_i m_i^q \,\frac{1}{D_q}\lVert \hat{q}_i-\bar{q}_i\rVert_1}
       {\sum_i m_i^q},
\end{equation}
where $i$ indexes tokens and acquisitions, $m_i^q$ denotes the corresponding observation mask ($m_k^p$ for $\Cenc{z}$, $o_{k,j}^p$ for $\Cenc{v}$; Sec.~\ref{sec:latent}), and $D_q$ is the token dimension, so each stream's error is averaged over its feature dimensions. Both streams are weighted equally, $\lambda_{\Cenc{z}}=\lambda_{\Cenc{v}}=1$; with the per-dimension average, the two terms are of similar magnitude throughout training (at the end of training, the teacher-forced loss is 0.49 for $\Cenc{z}$ and 0.55 for $\Cenc{v}$).



\paragraph{Training.} A predictor trained only with teacher forcing sees ground-truth inputs and is never exposed to its own errors, which then compound during rollout. Following V-JEPA~2-AC \cite{assran2025vjepa2}, we add rollout passes where the first pass is teacher-forced. Each subsequent pass keeps the $K_c$ observed context steps but replaces every later input state by the prediction of the previous pass, and is supervised on the target steps with $\mathcal{L}_{\text{step}}$. Each pass is a single block-causal forward pass that processes all steps in parallel, as in teacher forcing; the passes run sequentially, since each takes the previous pass's predictions as input. After $n$ passes, the prediction for step $K_c{+}j$ has been conditioned on its own outputs up to $\min(j{-}1, n{-}1)$ levels deep, at a cost of $n$ forward passes rather than the $K_f$ sequential steps of a full autoregressive rollout. The number of passes follows a curriculum of $2 \to 3 \to 4$ over thirds of training. Training on the model's own rollouts is essential for long horizons (Sec.~\ref{sec:ablations}). The total loss is

\begin{equation}
  \mathcal{L} = \mathcal{L}_{\text{step}}^{\text{tf}} + \frac{1}{n-1} \sum_{i=2}^{n} \mathcal{L}_{\text{step}}^{(i)}
\end{equation}
The implementation details, including the optimizer, schedule, and initialization, are given in supplementary Sec.~\ref{sec:impl}.

\subsection{Decoding}
\label{sec:decoding}

Forecasts are decoded into images by the EO-VAE decoder, $\hat{x}_k^j = \Cenc{D}(\Cenc{\hat{v}_k^j})$. With the predictor fixed, we fine-tune the decoder to reconstruct the future frames of training cubes from the predictor's rollout forecasts, adapting it to predicted tokens, whose distribution can differ from that of real ones; the loss is the EO-VAE reconstruction loss (masked $\ell_1$ in normalized band space plus an NDVI term over vegetated pixels). This step uses no test data and adds no parameters or inference cost. To isolate the predictor, all ablations instead decode with the adapted EO-VAE decoder before alignment, shared by every variant.

\subsection{Long-Horizon Inference}
\label{sec:longhorizon}
The predictor is trained on 150-day windows ($K_c{=}5$ context and $K_f{=}10$ forecast steps), while the long-horizon tracks span 200 and 700 days. We generate these forecasts through \emph{chunked rollout}. More precisely, the first chunk uses the last $K_c$ observed steps, and each subsequent chunk uses the last $K_c$ predicted steps from the previous chunk, for both $\Cenc{\hat{z}}$ and $\Cenc{\hat{v}}$. Thus, no observed imagery is reintroduced during rollout. Weather advances with each chunk, while the Tessera prior remains fixed to the year preceding the first observation. It is also important to clarify that we use observed ERA5-Land weather throughout, making this a hindcast of surface responses under known weather conditions, consistent with the EarthNet protocol, rather than a forecast of future weather. All long-horizon results are the raw rollout; no blending with persistence or climatology is applied.

\providecommand{\tbd}{\textcolor{red}{\textbf{TBD}}}

\section{Experiments}
\label{sec:experiments}

\subsection{Setup}
\label{sec:setup}

\begin{table*}[t]
\centering
\caption{\textbf{Extreme Summer and Seasonal benchmarks.} Extreme Summer evaluates 30-frame windows around summer extremes; Seasonal evaluates matched pairs from the same location under different weather conditions. Extreme metrics: ENS, P-MAE, N-MAE, trough NDVI MAE (TN-MAE), and drop-amplitude error (DAE), reported overall and by severity. Seasonal metrics: ENS, P-MAE, N-MAE, divergence-response ratio (DRR, best at 1), direction hit rate (DHR), and Spearman correlation of predicted and observed pair differences (PDC). Existing model results are reported as in EO-WM. EO-JEPA uses seed 0; over 3 seeds, Extreme ENS is $0.272 \pm 0.001$ and DAE is $0.228 \pm 0.010$, while Seasonal DHR is $0.756 \pm 0.008$ and PDC is $0.594 \pm 0.014$. Bold: best; underline: second best.}
\label{tab:combined}
\setlength{\tabcolsep}{3pt}
\resizebox{\linewidth}{!}{%
\begin{tabular}{l|ccc|cc|cc|cc|ccc|ccc}
\toprule
& \multicolumn{9}{c|}{Extreme Summer} & \multicolumn{6}{c}{Seasonal} \\
\cmidrule(lr){2-10}\cmidrule(lr){11-16}
\multirow{2}{*}{Model} & \multicolumn{3}{c|}{Overall} & \multicolumn{2}{c|}{Low} & \multicolumn{2}{c|}{Mid} & \multicolumn{2}{c|}{High} & \multicolumn{3}{c|}{Accuracy} & \multicolumn{3}{c}{Weather response} \\
& ENS$\uparrow$ & P-MAE$\downarrow$ & N-MAE$\downarrow$ & TN-MAE$\downarrow$ & DAE$\downarrow$ & TN-MAE$\downarrow$ & DAE$\downarrow$ & TN-MAE$\downarrow$ & DAE$\downarrow$ & ENS$\uparrow$ & P-MAE$\downarrow$ & N-MAE$\downarrow$ & DRR$\rightarrow\!1$ & DHR$\uparrow$ & PDC$\uparrow$ \\
\midrule
\multicolumn{16}{l}{\textit{Deterministic models}} \\
SimVP \cite{gao2022simvp} & \cellcolor{cyan!13}0.118 & \cellcolor{cyan!24}0.069 & \cellcolor{cyan!26}0.185 & \cellcolor{cyan!23}0.214 & \cellcolor{cyan!13}0.286 & \cellcolor{cyan!23}0.228 & \cellcolor{cyan!24}0.284 & \cellcolor{cyan!23}0.225 & \cellcolor{cyan!21}0.290 & \cellcolor{cyan!24}0.210 & \cellcolor{cyan!28}0.048 & \cellcolor{cyan!28}0.159 & \cellcolor{cyan!19}1.447 & \cellcolor{cyan!10}0.494 & \cellcolor{cyan!18}0.161 \\
PhyDNet \cite{guen2020phydnet} & \cellcolor{cyan!13}0.116 & \cellcolor{cyan!17}0.092 & \cellcolor{cyan!17}0.264 & \cellcolor{cyan!17}0.273 & \cellcolor{cyan!5}0.332 & \cellcolor{cyan!17}0.276 & \cellcolor{cyan!7}0.346 & \cellcolor{cyan!20}0.246 & \cellcolor{cyan!6}0.346 & \cellcolor{cyan!8}0.063 & \cellcolor{cyan!11}0.112 & \cellcolor{cyan!12}0.323 & \cellcolor{cyan!16}1.528 & \cellcolor{cyan!11}0.505 & \cellcolor{cyan!13}0.036 \\
PredRNN \cite{wang2017} & \cellcolor{cyan!5}0.059 & \cellcolor{cyan!5}0.136 & \cellcolor{cyan!5}0.376 & \cellcolor{cyan!5}0.378 & \cellcolor{cyan!8}0.317 & \cellcolor{cyan!5}0.381 & \cellcolor{cyan!13}0.323 & \cellcolor{cyan!5}0.372 & \cellcolor{cyan!14}0.315 & \cellcolor{cyan!5}0.031 & \cellcolor{cyan!5}0.133 & \cellcolor{cyan!5}0.394 & \cellcolor{cyan!5}0.162 & \cellcolor{cyan!5}0.444 & \cellcolor{cyan!5}$-$0.150 \\
PredRNN v2 \cite{wang2022predrnn} & \cellcolor{cyan!17}0.143 & \cellcolor{cyan!25}0.066 & \cellcolor{cyan!20}0.236 & \cellcolor{cyan!18}0.264 & \cellcolor{cyan!11}0.296 & \cellcolor{cyan!16}0.287 & \cellcolor{cyan!17}0.310 & \cellcolor{cyan!17}0.275 & \cellcolor{cyan!16}0.307 & \cellcolor{cyan!21}0.187 & \cellcolor{cyan!12}0.107 & \cellcolor{cyan!8}0.361 & \cellcolor{cyan!13}0.396 & \cellcolor{cyan!11}0.505 & \cellcolor{cyan!16}0.123 \\
TAU \cite{tan2023tau} & \cellcolor{cyan!16}0.135 & \cellcolor{cyan!21}0.078 & \cellcolor{cyan!21}0.229 & \cellcolor{cyan!20}0.246 & \cellcolor{cyan!7}0.323 & \cellcolor{cyan!20}0.255 & \cellcolor{cyan!5}0.352 & \cellcolor{cyan!20}0.244 & \cellcolor{cyan!5}0.349 & \cellcolor{cyan!20}0.173 & \cellcolor{cyan!28}0.048 & \cellcolor{cyan!28}0.151 & \cellcolor{cyan!18}0.518 & \cellcolor{cyan!10}0.495 & \cellcolor{cyan!17}0.139 \\
Earthformer \cite{gao2022earthformer} & \cellcolor{cyan!33}\underline{0.254} & \cellcolor{cyan!33}0.035 & \cellcolor{cyan!34}\underline{0.110} & \cellcolor{cyan!33}\underline{0.127} & \cellcolor{cyan!24}0.223 & \cellcolor{cyan!34}0.135 & \cellcolor{cyan!25}0.280 & \cellcolor{cyan!34}0.134 & \cellcolor{cyan!16}0.308 & \cellcolor{cyan!32}0.288 & \cellcolor{cyan!33}0.028 & \cellcolor{cyan!32}0.111 & \cellcolor{cyan!22}0.630 & \cellcolor{cyan!16}0.555 & \cellcolor{cyan!18}0.181 \\
\midrule
\multicolumn{16}{l}{\textit{Generative models}} \\
Wan2.1-Fun-InP* \cite{wan2025} & \cellcolor{cyan!25}0.203 & \cellcolor{cyan!31}0.043 & \cellcolor{cyan!31}0.135 & \cellcolor{cyan!32}0.137 & \cellcolor{cyan!25}\underline{0.218} & \cellcolor{cyan!33}0.145 & \cellcolor{cyan!32}\underline{0.254} & \cellcolor{cyan!33}0.143 & \cellcolor{cyan!26}\underline{0.269} & \cellcolor{cyan!28}0.254 & \cellcolor{cyan!32}0.033 & \cellcolor{cyan!31}0.121 & \cellcolor{cyan!25}\underline{0.707} & \cellcolor{cyan!20}0.605 & \cellcolor{cyan!23}0.281 \\
Latte \cite{ma2024latte} & \cellcolor{cyan!18}0.148 & \cellcolor{cyan!27}0.058 & \cellcolor{cyan!24}0.198 & \cellcolor{cyan!26}0.193 & \cellcolor{cyan!22}0.234 & \cellcolor{cyan!27}0.197 & \cellcolor{cyan!23}0.287 & \cellcolor{cyan!26}0.202 & \cellcolor{cyan!18}0.302 & \cellcolor{cyan!23}0.205 & \cellcolor{cyan!28}0.045 & \cellcolor{cyan!27}0.168 & \cellcolor{cyan!21}1.402 & \cellcolor{cyan!16}0.561 & \cellcolor{cyan!19}0.188 \\
EO-WM \cite{luo2026eowm} & \cellcolor{cyan!33}0.254 & \cellcolor{cyan!34}\underline{0.033} & \cellcolor{cyan!34}0.111 & \cellcolor{cyan!33}0.127 & \cellcolor{cyan!25}0.219 & \cellcolor{cyan!35}\underline{0.130} & \cellcolor{cyan!35}\textbf{0.243} & \cellcolor{cyan!35}\underline{0.128} & \cellcolor{cyan!35}\textbf{0.237} & \cellcolor{cyan!33}\underline{0.302} & \cellcolor{cyan!34}\underline{0.022} & \cellcolor{cyan!33}\underline{0.101} & \cellcolor{cyan!24}0.693 & \cellcolor{cyan!25}\underline{0.652} & \cellcolor{cyan!23}\underline{0.294} \\
\midrule
\multicolumn{16}{l}{\textit{Latent world model}} \\
\textbf{EO-JEPA (ours)} & \cellcolor{cyan!35}\textbf{0.271} & \cellcolor{cyan!35}\textbf{0.029} & \cellcolor{cyan!35}\textbf{0.102} & \cellcolor{cyan!35}\textbf{0.111} & \cellcolor{cyan!35}\textbf{0.163} & \cellcolor{cyan!35}\textbf{0.127} & \cellcolor{cyan!29}0.263 & \cellcolor{cyan!35}\textbf{0.124} & \cellcolor{cyan!26}0.270 & \cellcolor{cyan!35}\textbf{0.317} & \cellcolor{cyan!35}\textbf{0.020} & \cellcolor{cyan!35}\textbf{0.082} & \cellcolor{cyan!27}\textbf{0.777} & \cellcolor{cyan!35}\textbf{0.760} & \cellcolor{cyan!35}\textbf{0.587} \\
\midrule
\textcolor{gray}{Persistence} & \textcolor{gray}{0.129} & \textcolor{gray}{0.061} & \textcolor{gray}{0.173} & \textcolor{gray}{0.177} & \textcolor{gray}{0.282} & \textcolor{gray}{0.186} & \textcolor{gray}{0.427} & \textcolor{gray}{0.197} & \textcolor{gray}{0.464} & \textcolor{gray}{0.277} & \textcolor{gray}{0.030} & \textcolor{gray}{0.126} & \textcolor{gray}{0.651} & \textcolor{gray}{0.536} & \textcolor{gray}{0.216} \\
\bottomrule
\end{tabular}}
\end{table*}

\paragraph{Benchmarks.}
\emph{EarthNet2021}~\cite{requena2021earthnet} provides the IID and OOD test tracks (10 context frames, 20 target frames, i.e.\ 100 days) and two long-horizon tracks: \emph{Extreme} (20 context, 40 target frames, 200 days, during the 2018 European drought) and \emph{Seasonal} (70 context, 140 target frames, 700 days); each long track contains 4{,}000 cubes. We report the official EarthNetScore (ENS), the harmonic mean of its value (MAD), trend (OLS), distribution (EMD) and perceptual (SSIM) components.
The \emph{EO-WM} benchmarks~\cite{luo2026eowm} test the weather response directly. \emph{Extreme Summer} has 1{,}440 windows around summer extremes in three severity bins and reports, besides ENS, pixel and NDVI errors, the NDVI error at the trough (TN-MAE) and the error in the size of the drop (DAE). \emph{Seasonal} has 422 matched pairs, the same location in two years with different weather: the direction hit rate (DHR) counts how often the predicted difference between the years has the observed sign, the pair-difference correlation (PDC) is the Spearman correlation of predicted and observed differences, and the divergence-response ratio (DRR) compares their magnitudes (1 is ideal).

\subsection{Comparative Evaluation}
\label{sec:main_results}
\begin{figure*}[t]
\centering
\includegraphics[width=\linewidth]{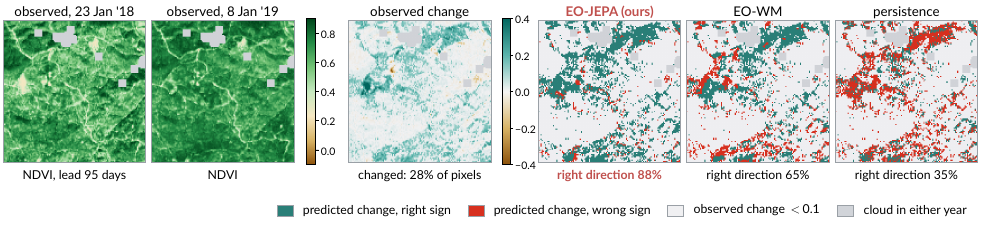}
\caption{\textbf{EO-JEPA predicts the direction of year-to-year vegetation change.}
Observed NDVI at the same Seasonal benchmark location and seasonal position in two different years, with a 95-day forecast lead.
The example is selected from the 79 vegetated pairs with the strongest year-to-year contrasts (top quartile of observed NDVI divergence, with a frame at least 90\% cloud-free in both years at a lead of 50 days or more) and lies at the 75th percentile of EO-JEPA's advantage over EO-WM.
For pixels with an observed absolute NDVI change of at least 0.1, the prediction maps indicate whether each model correctly predicts the direction of change.
EO-JEPA achieves 88\% directional accuracy in this example, compared with 65\% for EO-WM and 35\% for persistence.
Across all 79 pairs, the corresponding accuracies are 74\%, 65\%, and 57\%, respectively.
All models underestimate the magnitude of change.}
\label{fig:change}
\end{figure*}

\paragraph{Long-horizon forecasting} 
Tab.~\ref{tab:long} reports the long-horizon results, with forecasts generated through raw chunked rollout. On the 200-day Extreme track, EO-JEPA achieves an ENS of 0.282, exceeding SGConvLSTM by 0.8 points and attaining the best or tied-best score on every component. On the 700-day Seasonal track, it reaches 0.263, outperforming ConvLSTM, the strongest learned model, by 4.4 points. It also surpasses context-mean persistence on the trend, value, and distribution components, with OLS, MAD, and EMD scores of 0.392, 0.234, and 0.228, compared with 0.385, 0.233, and 0.203. Its lower SSIM score of 0.252, compared with 0.318, reflects a limitation of latent-space forecasting: predicting future representations does not explicitly optimize pixel-level structural similarity. Two weather-blind baselines score higher in ENS at 700 days, context-mean persistence (0.268) and the training-free seasonal anchor (0.275), as this track rewards keeping the seasonal cycle; neither responds to weather (PDC 0.216 for persistence versus 0.587; Tab.~\ref{tab:combined}).

Fig.~\ref{fig:lead} traces forecasts across successive 100-day chunks, each conditioned only on predictions from the preceding chunk. EO-JEPA follows the observed mean NDVI through the 2018 drought and across both seasonal cycles, with mean absolute trajectory deviations of 0.023 and 0.017 NDVI, respectively. Its error also stabilizes after the first year rather than continuing to diverge. In contrast, the published models drift over time. SGConvLSTM and SGEDConvLSTM underestimate NDVI in the second year, with errors of 0.172 and 0.213 compared with 0.117 for EO-JEPA, while the ConvLSTM of Diaconu et al. underestimates NDVI by about 0.10 throughout the Extreme track. EO-JEPA's pixel round trip also drifts high by 0.05 to 0.06 and accumulates error faster than the latent rollout.

\paragraph{Extreme events.}

On Extreme Summer (Tab.~\ref{tab:combined}), EO-JEPA achieves the highest ENS (0.271, or $0.272 \pm 0.001$ over three seeds, compared with 0.254 for both EO-WM and Earthformer) and the lowest pixel and NDVI errors. Its trough-NDVI error is the lowest across all severity bins, indicating more accurate predictions of minimum vegetation response than any baseline. Its drop-amplitude error is also the lowest for low-severity extremes (0.163 versus 0.218 for the best baseline), but exceeds EO-WM's on mid- and high-severity events (0.263 versus 0.243 and 0.270 versus 0.237, respectively). EO-JEPA thus captures the timing and depth of the troughs but underestimates the largest drops, consistent with its damped year-to-year differences on the Seasonal benchmark. Aligned on the observed trough, EO-JEPA follows the decline and recovery more closely than EO-WM in all three tiers (supplementary Fig.~\ref{fig:extreme_traj}). 


\paragraph{Response to weather.}
The Seasonal benchmark (Tab.~\ref{tab:combined}) evaluates whether a model captures differences at the same location across years with different weather conditions. EO-JEPA predicts the direction of these differences for 76\% of target steps, achieving a DHR of 0.760 compared with 0.652 for EO-WM. Its predicted differences also correlate better with observations than those of any other model, with a PDC of 0.587 versus 0.294 for EO-WM (Fig.~\ref{fig:hook}). At the pixel level, EO-JEPA also more often predicts the direction of change between the two years (Fig.~\ref{fig:change}). Its divergence ratio of 0.777, compared with 0.693 for EO-WM, indicates that it captures the direction and ordering of weather-driven changes but underestimates their magnitude, consistent with the Extreme benchmark. Inference controls further link this response to the weather input (Tab.~\ref{tab:main_controls}). Withholding weather reduces EO-JEPA's PDC from 0.587 to 0.432, while substituting weather from another location reduces it to 0.255. By contrast, its EarthNet2021 IID score changes by less than one point, from 0.329 to 0.321. The same controls change EO-WM's PDC by at most 0.020 and its IID score by at most 0.2 points, suggesting that its year-to-year response is driven primarily by its context frames rather than weather inputs. Finally, rolling out through pixels instead of latents reduces PDC to 0.511 and the 700-day ENS from 0.263 to 0.199. 
\begin{table}[t]
\centering
\caption{\textbf{Weather and decoding controls.} Ablations: null or mismatched-location weather; pixel round trip (decode and re-encode per step). EO-WM uses released weights (DHR 0.652, PDC 0.294). --: not run. Results in supplementary Tabs.~\ref{tab:controls} and~\ref{tab:weather_models}.}
\label{tab:main_controls}
\scriptsize
\setlength{\tabcolsep}{2pt}
\resizebox{\columnwidth}{!}{%
\begin{tabular}{l|cc|cc|cc}
\toprule
\multirow{2}{*}{Variant} & IID & Extreme & \multicolumn{2}{c|}{Seasonal} & \multicolumn{2}{c}{Long horizon} \\
 & ENS$\uparrow$ & ENS$\uparrow$ & DHR$\uparrow$ & PDC$\uparrow$ & 200\,d$\uparrow$ & 700\,d$\uparrow$ \\
\midrule
\multicolumn{7}{l}{\textit{EO-JEPA}} \\
As given & \cellcolor{cyan!35}0.329 & \cellcolor{cyan!35}0.271 & \cellcolor{cyan!35}0.760 & \cellcolor{cyan!35}0.587 & \cellcolor{cyan!35}0.282 & \cellcolor{cyan!35}0.263 \\
Weather withheld & \cellcolor{cyan!30}0.321 & \cellcolor{cyan!20}0.260 & \cellcolor{cyan!20}0.662 & \cellcolor{cyan!25}0.432 & \cellcolor{cyan!25}0.272 & \cellcolor{cyan!35}0.258 \\
Another location & \cellcolor{cyan!10}0.292 & \cellcolor{cyan!20}0.259 & \cellcolor{cyan!10}0.573 & \cellcolor{cyan!10}0.255 & \cellcolor{cyan!25}0.272 & \cellcolor{cyan!30}0.244 \\
Pixel round trip & \cellcolor{cyan!20}0.310 & \cellcolor{cyan!15}0.257 & \cellcolor{cyan!35}0.751 & \cellcolor{cyan!30}0.511 & \cellcolor{cyan!10}0.254 & \cellcolor{cyan!10}0.199 \\
\midrule
\multicolumn{7}{l}{\textit{EO-WM}} \\
As given & \cellcolor{cyan!35}0.317 & \cellcolor{cyan!35}0.253 & \cellcolor{cyan!35}0.650 & \cellcolor{cyan!35}0.285 & -- & -- \\
Weather withheld & \cellcolor{cyan!25}0.316 & \cellcolor{cyan!25}0.252 & \cellcolor{cyan!25}0.636 & \cellcolor{cyan!25}0.281 & -- & -- \\
Another location & \cellcolor{cyan!15}0.315 & \cellcolor{cyan!25}0.252 & \cellcolor{cyan!20}0.634 & \cellcolor{cyan!15}0.265 & -- & -- \\
\bottomrule
\end{tabular}}
\end{table}


\subsection{Ablation Study}
\label{sec:ablations}

Our ablations (Tab.~\ref{tab:main_ablation}; detailed results in Tab.~\ref{tab:ablation} supplementary, with discussion in Sec.~\ref{supp:ablations}) identify four factors underlying EO-JEPA's performance.

\paragraph{Semantic representations.} Removing the DOFA stream leaves EarthNet2021 performance largely unchanged but reduces Extreme Summer ENS by 0.4 points and PDC by 0.033; both its use as context and as a prediction target contribute to the weather response.

\paragraph{Rollout training.} Teacher-forced training reduces ENS by 1.6 points on Extreme Summer, 1.5 at 200 days and 0.8 at 700 days, so training on the model's own predictions is essential for chunked forecasting over hundreds of days.

\paragraph{Spatiotemporal mixing.} Removing spatial and temporal interactions reduces ENS by 8.6 points on Extreme Summer and 6.9 on the 700-day track, whereas a matched-size ConvLSTM reproduces the drop amplitudes and the year-to-year response at least as well but has lower ENS under extremes and over long rollouts (1.7 and 2.8 points); joint reasoning is essential, while the specific architecture is less decisive.
\begin{table}[t]
\centering
\caption{\textbf{Component ablations} (more in Tab.~\ref{tab:ablation}; shared, unaligned decoder). Three-seed means, except the last three rows (seed 0) and two seeds for the long-horizon columns of the first row.}
\label{tab:main_ablation}
\scriptsize
\setlength{\tabcolsep}{3pt}
\resizebox{\columnwidth}{!}{%
\begin{tabular}{l|c|c|c|cc}
\toprule
\multirow{2}{*}{Variant} & EN21 & Extreme & Seasonal & \multicolumn{2}{c}{Long horizon} \\
 & IID$\uparrow$ & ENS$\uparrow$ & PDC$\uparrow$ & 200\,d$\uparrow$ & 700\,d$\uparrow$ \\
\midrule
V-JEPA~2 init. (EO-JEPA) & \cellcolor{cyan!30}0.305 & \cellcolor{cyan!20}0.245 & \cellcolor{cyan!25}0.548 & \cellcolor{cyan!20}0.243 & \cellcolor{cyan!15}0.214 \\
Random init. & \cellcolor{cyan!35}0.307 & \cellcolor{cyan!30}0.253 & \cellcolor{cyan!30}0.557 & \cellcolor{cyan!25}0.249 & \cellcolor{cyan!20}0.225 \\
\midrule
No semantic stream $z$ & \cellcolor{cyan!35}0.307 & \cellcolor{cyan!25}0.249 & \cellcolor{cyan!15}0.524 & \cellcolor{cyan!20}0.246 & \cellcolor{cyan!15}0.231 \\
ConvLSTM, matched size & \cellcolor{cyan!25}0.303 & \cellcolor{cyan!15}0.236 & \cellcolor{cyan!35}0.592 & \cellcolor{cyan!15}0.239 & \cellcolor{cyan!10}0.197 \\
Per-token head, no mixing & \cellcolor{cyan!10}0.268 & \cellcolor{cyan!10}0.167 & \cellcolor{cyan!10}0.343 & \cellcolor{cyan!10}0.192 & \cellcolor{cyan!10}0.156 \\
\midrule
Teacher forcing only & \cellcolor{cyan!15}0.300 & \cellcolor{cyan!15}0.237 & \cellcolor{cyan!30}0.556 & \cellcolor{cyan!15}0.234 & \cellcolor{cyan!15}0.217 \\
Pixel round trip & \cellcolor{cyan!10}0.277 & \cellcolor{cyan!10}0.214 & \cellcolor{cyan!15}0.444 & \cellcolor{cyan!10}0.153 & \cellcolor{cyan!10}0.104 \\
\bottomrule
\end{tabular}}
\end{table}

\paragraph{Initialization.} V-JEPA~2 weights do not improve accuracy over random initialization, although they yield stronger weather-response metrics with aligned decoders, so the gains arise from learning latent next-state dynamics rather than from pretrained predictor weights.

\section{Conclusion}
\label{sec:conclusion}

We introduce EO-JEPA, a latent world model for forecasting weather-driven land-surface dynamics. It represents the land surface as a latent state and uses meteorology to drive its evolution. Built from two frozen foundation-model encoders and a 25M-parameter predictor, EO-JEPA achieves the strongest results among learned models on Extreme Summer and the 200- and 700-day tracks, doubles the correlation between predicted and observed year-to-year differences, and remains competitive at the standard 100-day horizon. Controlled experiments attribute these gains to latent-space rollout and explicit weather conditioning: withholding or shuffling the weather removes the model's response to conditions while barely affecting EO-WM, whose response comes from its context frames. Notably, withholding weather changes the EarthNet2021 score by less than one point but removes a quarter of the paired response, showing that accuracy metrics alone cannot establish whether a forecaster responds to weather. We therefore recommend reporting paired-response metrics alongside forecasting scores.

\paragraph{Limitations.}
All forecasts are driven by the observed ERA5-Land weather of the forecast period, as in the benchmarks; they measure how well the land surface follows known weather, not operational forecast skill. The model is deterministic, which contributes to its lower SSIM on the 700-day track, where context-mean persistence and the seasonal anchor remain higher in ENS, and may explain why it underestimates the depth of the most severe vegetation drops. Accuracy at 100 days remains below Earthformer. Most inference cost is spent in the EO-VAE decoder, and the best scores use a decoder aligned to the predictor's outputs; all benchmarks cover Europe at 20\,m resolution, where frozen encoders bound what the state can represent.


{
    \small
    \bibliographystyle{plainnat}
    \bibliography{main}
}

\clearpage
\setcounter{page}{1}
\maketitlesupplementary
\setcounter{section}{0}\renewcommand{\thesection}{S\arabic{section}}
\setcounter{table}{0}\renewcommand{\thetable}{S\arabic{table}}
\setcounter{figure}{0}\renewcommand{\thefigure}{S\arabic{figure}}
\setcounter{equation}{0}\renewcommand{\theequation}{S\arabic{equation}}

This supplementary material gives the implementation details (Sec. \ref{sec:impl}), extended related work (Sec. \ref{supp:related}) and further design details (Sec. \ref{supp:design}), the results on the standard EarthNet2021 horizon (Sec. \ref{supp:en21}), the inference-time weather controls (Sec. \ref{supp:weather}), the pixel round trip (Sec. \ref{supp:rollout}), the results by land cover (Sec. \ref{supp:landcover}) and the component ablations (Sec. \ref{supp:ablations}).

\section{Implementation details}
\label{sec:impl}

\begin{figure*}[t]
\centering
\includegraphics[width=\linewidth]{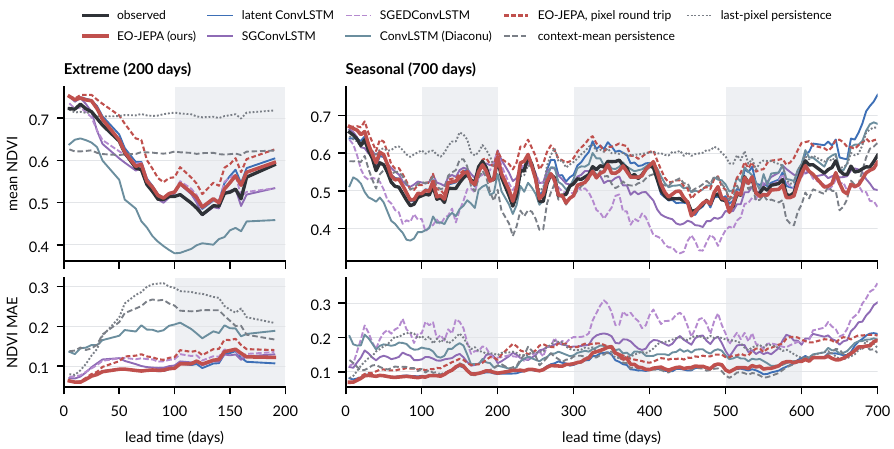}
\caption{\textbf{Fig.~\ref{fig:lead} with every source}, adding the latent ConvLSTM (a predictor of matched size trained in our latent space, aligned decoder) and last-pixel persistence. Same cubes, pixels and smoothing as Fig.~\ref{fig:lead}.}
\label{fig:supp_lead}
\end{figure*}

\paragraph{Implementation and compute.}
Every stage runs on a single NVIDIA A100 (40\,GB). First, EO-VAE is adapted to the EarthNet2021 training imagery (20k steps, batch size 8; 3.6\,h), and its tokens of all training frames are cached once; DOFA features are computed on the fly. Second, the V-JEPA~2 predictor \cite{assran2025vjepa2} is adapted to the same imagery by continued self-supervised V-JEPA pretraining (30k steps; 10.3\,h). Third, it is trained as the action-conditioned predictor of Sec. \ref{sec:training} for 30k steps with batch size 8 (8.0\,h, 11\,GB of GPU memory), using AdamW with a learning rate of $10^{-4}$ ($3\times$ for newly added modules) and weight decay 0.04, 1k warm-up steps and a cosine decay to $10^{-6}$, gradient clipping at 1.0, condition dropout 0.15 and the $2{\to}3{\to}4$ rollout curriculum; the checkpoint with the lowest validation loss on 2\% held-out training cubes is kept. Finally, the decoder is aligned on rollout forecasts of the training cubes (58.5k steps; 7.1\,h). A complete model thus takes about 29 GPU-hours, and each ablation, which retrains only the predictor, 8--11 GPU-hours on a dedicated GPU. Of its 232M parameters, only the 25M of the predictor (and, during alignment, the 55M of the decoder) are trained; the 111M of DOFA and the 40M of the EO-VAE encoder stay frozen.
At inference, a 100-day forecast costs 18.4~TFLOPs and 0.31\,s per cube at batch size 8 (0.54\,s at batch size 1). Most of it is spent rendering pixels: the EO-VAE decoder accounts for 12.5~TFLOPs and the latent rollout of the predictor for 2.7~TFLOPs. Earthformer, which forecasts in pixel space, needs 1.8~TFLOPs and 0.08\,s per cube. Long-horizon forecasts chain the same computation once per 100-day chunk, so their cost grows linearly with the horizon.
Unless stated otherwise, results are for seed 0 with the aligned decoder (Sec. \ref{sec:decoding}); we report the mean and standard deviation over three training seeds where available.

\paragraph{Baselines.}
Published results are taken as reported by their sources: EarthNet2021 baselines and long-horizon models \cite{requena2021earthnet,diaconu2022understanding,kladny2024enhanced}, Earthformer \cite{gao2022earthformer}, and the deterministic and generative video models evaluated by EO-WM \cite{luo2026eowm}. Persistence repeats the last cloud-free context pixel; on EarthNet2021 we also list the official context-mean persistence. Neither is ranked. The seasonal anchor of Tab.~\ref{tab:long} uses the 70 context frames (350 days) of the Seasonal track: for each target frame, it takes per pixel the nearest cloud-free context frame at the same position in the 73-frame year, searched within $\pm 3$ frames (15 days), with the circular neighbours for the three year positions that the context does not cover, and the cloud-free context mean where no frame is clear. It uses no model, training or weather, and the context-mean row computed by the same code reproduces the published score (0.268 ENS), which checks our scoring. The 100-day context of the Extreme track cannot support it. For the trajectory figures and the weather controls, we also run the long-horizon models of Kladny et al. \cite{kladny2024enhanced} and Diaconu et al. \cite{diaconu2022understanding} and EO-WM \cite{luo2026eowm} with their released weights and inference code. Our runs reproduce their reported scores to within 0.006 ENS (Diaconu et al.\ and the 200-day scores exactly; the 700-day scores of Kladny et al.\ are 0.006 higher because their land-cover masks are not available and we mask with the quality layer only) and EO-WM's PDC to within 0.01.

\paragraph{Train--test overlap.}
The EarthNet2021 test tracks, and the EO-WM windows drawn from them, are not fully disjoint from the training split: 11\% of the Extreme and 5\% of the Seasonal test cubes share their location with a training cube, mostly during the forecast period, which affects every model trained on this split. Excluding them changes EO-JEPA's long-horizon ENS by at most 0.0014 and the Seasonal-pair PDC from 0.587 to 0.565 (EO-WM: 0.285 to 0.318), and leaves every ranking unchanged.

\section{Extended related work}
\label{supp:related}

\paragraph{Physics-Informed Forecasting}
Physics-informed and hybrid models incorporate physical knowledge through governing-equation constraints \cite{raissi2019pinn}, differentiable process models \cite{reichstein2019deep,shen2023differentiable,kochkov2024neuralgcm}, or PDE-constrained latent dynamics \cite{guen2020phydnet, luo2026eowm}. Meanwhile, data-driven weather models increasingly rival numerical forecasts at medium range \cite{lam2023graphcast,bi2023pangu}, but predict atmospheric conditions rather than the land-surface responses captured by optical satellites. Vegetation dynamics lack closed-form governing equations at pixel resolution, making direct equation-based constraints difficult to formulate.

Rather than imposing equations, we encode physical structure through three inductive biases, namely, causal attention, lagged and cumulative meteorological forcing, and persistent land-state representations that retain prior-year land surface context. Then, we evaluate physical plausibility by testing whether predicted changes reflect the direction and magnitude of observed vegetation responses to contrasting weather conditions, beyond reconstruction accuracy alone.

\section{Further design details}
\label{supp:design}

\paragraph{Frozen encoders.}
We freeze both encoders to keep the representations stable during training, so the model learns how the Earth evolves rather than changing how it represents the input. In preliminary experiments, fine-tuning DOFA with an exponential-moving-average target, as in V-JEPA \cite{bardes2024vjepa}, caused the representations to drift and lose useful variation over time. Freezing the encoders therefore keeps learning focused on the dynamics.

\paragraph{Classifier-free guidance.}
{Condition dropout (Sec. \ref{sec:context}) also enables classifier-free guidance \cite{ho2022classifier} on the weather action, thereby: 

\begin{equation}
   \Cenc{\hat{s}} = \Cenc{\hat{s}}_\varnothing + \gamma\,(\Cenc{\hat{s}}_{\Cmet{a}} - \Cenc{\hat{s}}_\varnothing). 
\end{equation}
All reported forecasts use the conditional prediction ($\gamma{=}1$) unless stated otherwise.

\paragraph{Weather inputs and encoder.}
The action uses the 15 daily ERA5-Land variables of Tab.~\ref{tab:supp_weather}, sampled on a $10{\times}10$ grid at 10.24\,km centred on the cube (the 102.4\,km window of the EarthNet2021 mesoscale field), bilinearly interpolated to $40{\times}40$ and standardized per variable with training-set statistics; each window's series spans the 90 days before its first day and its 150 days. The encoder $\Cmet{g_\phi}$ (0.59M parameters) applies to every day a convolutional stem (three $3{\times}3$ convolutions with stride 2 and 32, 64 and 64 channels, each followed by GELU, then a linear map of the flattened $64{\times}5{\times}5$ map to 128 features), followed by two causal temporal convolutions over days (kernel 7, 128 channels, residual, GELU), which give each day's feature a receptive field of 13 days. Each action concatenates the mean and maximum of these features over its days (256 values) and maps them with a LayerNorm and a linear layer to the predictor width of 384.

\begin{table}[t]
\centering
\caption{\textbf{ERA5-Land inputs of the weather action}, before standardization. Daily values from the 00, 06, 12 and 18\,UTC fields; accumulated fluxes as daily totals.}
\label{tab:supp_weather}
\small
\setlength{\tabcolsep}{3pt}
\begin{tabularx}{\columnwidth}{@{}>{\raggedright\arraybackslash}Xll@{}}
\toprule
Variable & Daily value & Unit \\
\midrule
2\,m temperature & mean, min, max & $^\circ$C \\
Vapour-pressure deficit & mean, max & kPa \\
Surface solar radiation (down) & total & MJ\,m$^{-2}$ \\
Total precipitation & total & mm \\
Total evaporation (positive up) & total & mm \\
Surface pressure & mean & hPa \\
10\,m wind speed & mean & m\,s$^{-1}$ \\
Snow depth water equivalent & mean & m \\
Soil water (0--7, 7--28, 28--100, 100--289\,cm) & mean & m$^3$\,m$^{-3}$ \\
\bottomrule
\end{tabularx}
\end{table}

\paragraph{Elevation and date inputs.}
Both inputs are described in Sec.~\ref{sec:context} and are dropped independently with probability 0.15 during training, so they can be switched off at inference without retraining. Tab.~\ref{tab:supp_static} shows that switching them off, separately or together, changes ENS by at most 0.002 on every benchmark and PDC by at most 0.010, within the variation across training seeds ($\pm 0.014$); the model's skill does not depend on them.

\begin{table}[t]
\centering
\caption{\textbf{Elevation and date switched off at inference} (EO-JEPA, seed 0, aligned decoder). ENS: EarthNetScore; DHR / PDC: seasonal direction hit rate / pair-difference correlation.}
\label{tab:supp_static}
\setlength{\tabcolsep}{4pt}
\resizebox{\linewidth}{!}{%
\begin{tabular}{l|cccc|cc}
\toprule
Inputs & IID & OOD & Extreme & Seasonal & DHR & PDC \\
\midrule
All (as trained) & \cellcolor{cyan!35}0.3295 & \cellcolor{cyan!35}0.3203 & \cellcolor{cyan!35}0.2712 & \cellcolor{cyan!35}0.3171 & \cellcolor{cyan!30}0.760 & \cellcolor{cyan!35}0.587 \\
No elevation & \cellcolor{cyan!30}0.3292 & \cellcolor{cyan!30}0.3197 & \cellcolor{cyan!35}0.2712 & \cellcolor{cyan!30}0.3167 & \cellcolor{cyan!35}0.761 & \cellcolor{cyan!30}0.586 \\
No date & \cellcolor{cyan!20}0.3286 & \cellcolor{cyan!20}0.3194 & \cellcolor{cyan!15}0.2708 & \cellcolor{cyan!20}0.3158 & \cellcolor{cyan!15}0.754 & \cellcolor{cyan!20}0.578 \\
No elevation, no date & \cellcolor{cyan!15}0.3284 & \cellcolor{cyan!15}0.3189 & \cellcolor{cyan!15}0.2708 & \cellcolor{cyan!15}0.3153 & \cellcolor{cyan!20}0.755 & \cellcolor{cyan!15}0.577 \\
\bottomrule
\end{tabular}}
\end{table}

\section{Standard horizon}
\label{supp:en21}
\begin{figure*}
    \centering
\includegraphics{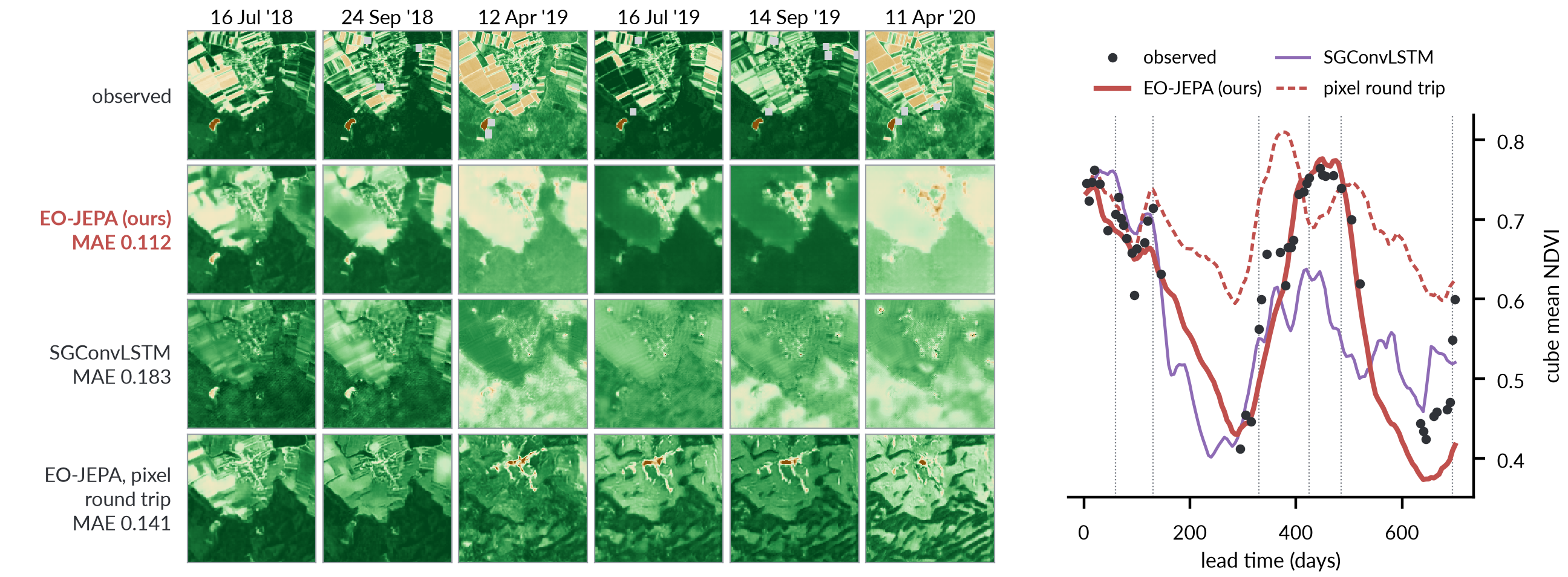}
    \caption{\textbf{One cube over 700 days.} NDVI maps of a Seasonal-track cube (tile 33TWN) at six cloud-free dates from July 2018 to April 2020 (left; grey: clouds in the observation), and the cube's mean NDVI over the full target period (right; observed: frames at least half cloud-free; forecasts: all pixels, 3-frame moving average). The cube is not hand-picked: among the 3{,}148 of 4{,}000 test cubes that have a cloud-free frame ($\geq$90\% clear pixels) in each sixth of the target period and are vegetated (mean NDVI $\geq$ 0.3), it is the one at the median of EO-JEPA's per-cube EarthNetScore (0.256; SGConvLSTM 0.186, pixel round trip 0.209). The NDVI MAE of each forecast on this cube is given next to it. EO-JEPA (aligned decoder) keeps the vegetation state and its seasonal timing over both years, including the decline in the second summer, while SGConvLSTM (released weights) settles too low in the second year and the pixel round trip of the same EO-JEPA model drifts high. EO-JEPA's maps are smoother than the observations, consistent with its lower perceptual (SSIM) score: field-level texture fades over the rollout.}
    \label{fig:supp_qual}
\end{figure*}
\begin{table*}[t]
\centering
\caption{\textbf{EarthNet2021 IID and OOD tracks} (official EarthNetScore and sub-scores; 10 context $\rightarrow$ 20 target frames). Published results as reported; EO-WM$^\dagger$: released weights, run by us. EO-JEPA: seed 0; three-seed mean: IID $0.330 \pm 0.001$, OOD $0.321 \pm 0.001$. Bold: best; underline: second best; gray: not ranked.}
\label{tab:en21}
\footnotesize
\setlength{\tabcolsep}{4pt}
\renewcommand{\arraystretch}{1.08}
\begin{tabular}{l|ccccc|ccccc}
\toprule
\multirow{2}{*}{Model} & \multicolumn{5}{c|}{IID} & \multicolumn{5}{c}{OOD} \\
 & ENS$\uparrow$ & MAD$\uparrow$ & OLS$\uparrow$ & EMD$\uparrow$ & SSIM$\uparrow$ & ENS$\uparrow$ & MAD$\uparrow$ & OLS$\uparrow$ & EMD$\uparrow$ & SSIM$\uparrow$ \\
\midrule
\multicolumn{11}{l}{\textit{Published models}} \\
Arcon \cite{requena2021earthnet}
& \cellcolor{cyan!12}0.280 & \cellcolor{cyan!12}0.241 & \cellcolor{cyan!12}0.322 & \cellcolor{cyan!12}0.226 & \cellcolor{cyan!12}0.386
& \cellcolor{cyan!12}0.266 & \cellcolor{cyan!12}0.231 & \cellcolor{cyan!12}0.309 & \cellcolor{cyan!12}0.218 & \cellcolor{cyan!12}0.343 \\

Channel-U-Net \cite{requena2021earthnet}
& \cellcolor{cyan!19}0.290 & \cellcolor{cyan!19}0.248 & \cellcolor{cyan!19}0.338 & \cellcolor{cyan!19}0.234 & \cellcolor{cyan!19}0.397
& \cellcolor{cyan!19}0.285 & \cellcolor{cyan!19}0.240 & \cellcolor{cyan!19}0.339 & \cellcolor{cyan!19}0.237 & \cellcolor{cyan!19}0.372 \\

EO-WM$^\dagger$ \cite{luo2026eowm}
& \cellcolor{cyan!23}0.317 & \cellcolor{cyan!23}0.259 & \cellcolor{cyan!23}0.350 & \cellcolor{cyan!23}0.256 & \cellcolor{cyan!23}0.504
& \cellcolor{cyan!23}0.307 & \cellcolor{cyan!23}0.247 & \cellcolor{cyan!23}0.349 & \cellcolor{cyan!23}0.259 & \cellcolor{cyan!23}0.446 \\

ConvLSTM \cite{diaconu2022understanding}
& \cellcolor{cyan!27}0.327 & \cellcolor{cyan!27}0.264 & \cellcolor{cyan!27}0.351 & \cellcolor{cyan!27}0.262 & \cellcolor{cyan!27}0.556
& \cellcolor{cyan!30}0.320 & \cellcolor{cyan!35}\underline{0.254} & \cellcolor{cyan!27}0.352 & \cellcolor{cyan!30}\underline{0.266} & \cellcolor{cyan!30}\underline{0.512} \\

Earthformer \cite{gao2022earthformer}
& \cellcolor{cyan!35}\textbf{0.343} & \cellcolor{cyan!35}\textbf{0.275} & \cellcolor{cyan!35}\textbf{0.362} & \cellcolor{cyan!35}\textbf{0.270} & \cellcolor{cyan!35}\textbf{0.635}
& \cellcolor{cyan!35}\textbf{0.325} & \cellcolor{cyan!30}0.254 & \cellcolor{cyan!35}\textbf{0.358} & \cellcolor{cyan!35}\textbf{0.273} & \cellcolor{cyan!35}\textbf{0.527} \\
\midrule
\multicolumn{11}{l}{\textit{Latent world model}} \\
\textbf{EO-JEPA (ours)}
& \cellcolor{cyan!30}\underline{0.329} & \cellcolor{cyan!30}\underline{0.267} & \cellcolor{cyan!30}\underline{0.354} & \cellcolor{cyan!30}\underline{0.265} & \cellcolor{cyan!30}\underline{0.558}
& \cellcolor{cyan!30}\underline{0.320} & \cellcolor{cyan!35}\textbf{0.255} & \cellcolor{cyan!30}\underline{0.353} & \cellcolor{cyan!27}0.266 & \cellcolor{cyan!27}0.506 \\
\midrule
\textcolor{gray}{Persistence \cite{requena2021earthnet}}
& \textcolor{gray}{0.271} & \textcolor{gray}{0.236} & \textcolor{gray}{0.324} & \textcolor{gray}{0.221} & \textcolor{gray}{0.342}
& \textcolor{gray}{0.265} & \textcolor{gray}{0.228} & \textcolor{gray}{0.324} & \textcolor{gray}{0.223} & \textcolor{gray}{0.318} \\
\bottomrule
\end{tabular}
\end{table*}

{On the EarthNet2021 IID and OOD tracks (Tab. \ref{tab:en21}) EO-JEPA scores 0.329 and 0.320 ($0.330 \pm 0.001$ and $0.321 \pm 0.001$ over three seeds), on par with the weather-conditioned ConvLSTM of Diaconu et al. \cite{diaconu2022understanding} and below Earthformer \cite{gao2022earthformer} (0.343 and 0.325). The latent formulation therefore keeps competitive accuracy over 100 days, where the context frames still dominate the forecast, and its advantage appears where the forecast must follow the weather: under extremes and over hundreds of days.

\section{Weather as the action}
\label{supp:weather}

\begin{table*}[t]
\centering
\caption{\textbf{Weather and decoding controls} on EO-JEPA (seed 0, aligned decoder), applied at inference to the same trained model. Withheld inputs are replaced by learned null embeddings. Another location assigns each window the daily weather of a different window using a fixed derangement. Pixel round trips decode and re-encode each predicted step or 100-day chunk before feedback. Persistence uses the last cloud-free pixel; EarthNet2021 results are as published. --: not applicable or not run.}
\label{tab:controls}
\scriptsize
\setlength{\tabcolsep}{3pt}
\begin{tabular}{l|cc|cc|ccc|cc}
\toprule
\multirow{2}{*}{Variant} & \multicolumn{2}{c|}{EarthNet2021} & \multicolumn{2}{c|}{Extreme Summer} & \multicolumn{3}{c|}{Seasonal pairs} & \multicolumn{2}{c}{Long horizon} \\
 & IID$\uparrow$ & OOD$\uparrow$ & ENS$\uparrow$ & DAE$\downarrow$ & ENS$\uparrow$ & DHR$\uparrow$ & PDC$\uparrow$ & 200\,d$\uparrow$ & 700\,d$\uparrow$ \\
\midrule 
Full model & \cellcolor{cyan!35}0.329 & \cellcolor{cyan!35}0.320 & \cellcolor{cyan!35}0.271 & \cellcolor{cyan!35}0.232 & \cellcolor{cyan!35}0.317 & \cellcolor{cyan!25}0.760 & \cellcolor{cyan!35}0.587 & \cellcolor{cyan!35}0.282 & \cellcolor{cyan!15}0.263 \\
Weather withheld & \cellcolor{cyan!30}0.321 & \cellcolor{cyan!30}0.315 & \cellcolor{cyan!20}0.260 & \cellcolor{cyan!15}0.274 & \cellcolor{cyan!30}0.313 & \cellcolor{cyan!20}0.662 & \cellcolor{cyan!25}0.432 & \cellcolor{cyan!25}0.272 & \cellcolor{cyan!10}0.258 \\
Another location & \cellcolor{cyan!10}0.292 & \cellcolor{cyan!10}0.284 & \cellcolor{cyan!15}0.259 & \cellcolor{cyan!30}0.223 & \cellcolor{cyan!10}0.288 & \cellcolor{cyan!10}0.573 & \cellcolor{cyan!15}0.255 & \cellcolor{cyan!25}0.272 & \cellcolor{cyan!10}0.244 \\
Tessera withheld & \cellcolor{cyan!30}0.328 & \cellcolor{cyan!30}0.318 & \cellcolor{cyan!30}0.269 & \cellcolor{cyan!35}0.217 & \cellcolor{cyan!30}0.316 & \cellcolor{cyan!30}0.768 & \cellcolor{cyan!30}0.584 & -- & -- \\
Weather and Tessera withheld & \cellcolor{cyan!25}0.317 & \cellcolor{cyan!25}0.311 & \cellcolor{cyan!15}0.259 & \cellcolor{cyan!20}0.260 & \cellcolor{cyan!25}0.311 & \cellcolor{cyan!20}0.663 & \cellcolor{cyan!20}0.383 & -- & -- \\
\midrule
\multicolumn{10}{l}{\textit{Decoding deferred to output vs.\ pixel-space rollout}} \\
Pixel round trip, every step & \cellcolor{cyan!15}0.310 & \cellcolor{cyan!15}0.301 & \cellcolor{cyan!10}0.257 & \cellcolor{cyan!25}0.242 & \cellcolor{cyan!15}0.301 & \cellcolor{cyan!15}0.751 & \cellcolor{cyan!20}0.511 & \cellcolor{cyan!10}0.254 & \cellcolor{cyan!10}0.199 \\
Pixel round trip, every chunk & -- & -- & -- & -- & -- & -- & -- & \cellcolor{cyan!35}0.282 & \cellcolor{cyan!20}0.269 \\
\midrule
\textcolor{gray}{Persistence} & \textcolor{gray}{0.271} & \textcolor{gray}{0.265} & \textcolor{gray}{0.129} & \textcolor{gray}{0.391} & \textcolor{gray}{0.277} & \textcolor{gray}{0.536} & \textcolor{gray}{0.216} & \textcolor{gray}{0.183} & \textcolor{gray}{0.210} \\
\bottomrule
\end{tabular}
\end{table*}

{Because the weather enters only through the action pathway and every conditioning input is dropped during training, we can test what the model does with the weather by changing it at inference, without retraining (Tab. \ref{tab:controls}, Fig. \ref{fig:response}).

\paragraph{The response is driven by the weather.}
{Withholding the weather (null action) costs 0.8 ENS points on IID, 1.1 on Extreme Summer, and 1.0 and 0.5 on the 200- and 700-day tracks, the largest loss among the individual inputs, but its effect on the weather response is far larger: PDC falls from 0.587 to 0.432, DHR from 0.760 to 0.662, and the drop-amplitude error grows from 0.232 to 0.274, so without weather the model forecasts a typical year. Giving every window the weather of a different location is more damaging still. PDC falls to 0.255, close to persistence (0.216), DHR to 0.573, and ENS drops by 3.8 points on IID, 1.2 on Extreme Summer and 2.9 on the Seasonal pairs: the model follows the weather it is given, even when that weather is wrong. The randomly initialized model behaves the same way (PDC 0.557 to 0.186 with the shared decoder). A model that ignored the weather would be unaffected by either change, and one that used it only as a generic seasonal cue would not lose its response when it is withheld. The remaining PDC without weather (0.432) reflects differences between the two years that are already visible in their context frames.

\begin{table}[t]
\centering
\caption{\textbf{Do the models use the weather?} Weather controls on EO-JEPA (seed 0, aligned decoder) and EO-WM (released weights, ensemble mean of 5). Withheld: EO-JEPA's null action; for EO-WM, its training-time weather dropout, which zeroes the daily weather anomalies and cumulative-stress features but keeps the tile's monthly climatology. Another location: all weather inputs of a different window (for EO-WM, including the climatology), with imagery, dates and captions unchanged.
}
\label{tab:weather_models}
\small
\setlength{\tabcolsep}{3pt}
\begin{tabular}{l|cc|ccc}
\toprule
\multirow{2}{*}{Weather input} & \multicolumn{2}{c|}{Extreme} & \multicolumn{3}{c}{Seasonal} \\
 & ENS$\uparrow$ & DAE$\downarrow$ & ENS$\uparrow$ & DHR$\uparrow$ & PDC$\uparrow$ \\
\midrule
\multicolumn{6}{l}{\textit{EO-JEPA}} \\
As given & \cellcolor{cyan!35}0.271 & \cellcolor{cyan!15}0.232 & \cellcolor{cyan!35}0.317 & \cellcolor{cyan!35}0.760 & \cellcolor{cyan!35}0.587 \\
Withheld & \cellcolor{cyan!20}0.260 & \cellcolor{cyan!10}0.274 & \cellcolor{cyan!30}0.313 & \cellcolor{cyan!20}0.662 & \cellcolor{cyan!25}0.432 \\
Another location & \cellcolor{cyan!15}0.259 & \cellcolor{cyan!25}0.223 & \cellcolor{cyan!15}0.288 & \cellcolor{cyan!15}0.573 & \cellcolor{cyan!15}0.255 \\
\midrule
\multicolumn{6}{l}{\textit{EO-WM}} \\
As given & \cellcolor{cyan!35}0.253 & \cellcolor{cyan!15}0.238 & \cellcolor{cyan!35}0.303 & \cellcolor{cyan!35}0.650 & \cellcolor{cyan!35}0.285 \\
Withheld & \cellcolor{cyan!20}0.252 & \cellcolor{cyan!15}0.238 & \cellcolor{cyan!20}0.302 & \cellcolor{cyan!20}0.636 & \cellcolor{cyan!25}0.281 \\
Another location & \cellcolor{cyan!20}0.252 & \cellcolor{cyan!25}0.236 & \cellcolor{cyan!15}0.301 & \cellcolor{cyan!15}0.634 & \cellcolor{cyan!15}0.265 \\
\bottomrule
\end{tabular}
\end{table}
\begin{table*}[t]
\centering
\caption{\textbf{Component ablations.} All rows decode with the adapted EO-VAE decoder before alignment, shared by every variant, so differences reflect the predictor. Rows are means over the $n$ training seeds shown (otherwise seed 0; ``long $n$'': seeds available for the two long-horizon columns). Unless stated, variants change one component of the randomly initialized model (``Random'' row): the semantic stream $z$ (DOFA) as both input and target, a 12-block transformer predictor and rollout training with $2{\to}3{\to}4$ passes.}
\label{tab:ablation}
\setlength{\tabcolsep}{4pt}
\resizebox{\linewidth}{!}{%
\begin{tabular}{l|cc|cc|ccc|cc}
\toprule
\multirow{2}{*}{Variant} & \multicolumn{2}{c|}{EarthNet2021} & \multicolumn{2}{c|}{Extreme Summer} & \multicolumn{3}{c|}{Seasonal pairs} & \multicolumn{2}{c}{Long horizon} \\
 & IID$\uparrow$ & OOD$\uparrow$ & ENS$\uparrow$ & DAE$\downarrow$ & ENS$\uparrow$ & DHR$\uparrow$ & PDC$\uparrow$ & 200\,d$\uparrow$ & 700\,d$\uparrow$ \\
\midrule
\multicolumn{10}{l}{\textit{Predictor initialization}} \\
V-JEPA~2 predictor (EO-JEPA) ($n{=}3$; long $n{=}2$) & \cellcolor{cyan!30}0.305 & \cellcolor{cyan!25}0.298 & \cellcolor{cyan!20}0.245 & \cellcolor{cyan!25}0.153 & \cellcolor{cyan!20}0.298 & \cellcolor{cyan!20}0.713 & \cellcolor{cyan!25}0.548 & \cellcolor{cyan!20}0.243 & \cellcolor{cyan!15}0.214 \\
Random ($n{=}3$) & \cellcolor{cyan!35}0.307 & \cellcolor{cyan!35}0.301 & \cellcolor{cyan!30}0.253 & \cellcolor{cyan!20}0.170 & \cellcolor{cyan!30}0.300 & \cellcolor{cyan!30}0.725 & \cellcolor{cyan!30}0.557 & \cellcolor{cyan!25}0.249 & \cellcolor{cyan!20}0.225 \\
\midrule
\multicolumn{10}{l}{\textit{Latent state (random init.)}} \\
No semantic stream $z$ ($n{=}3$) & \cellcolor{cyan!35}0.307 & \cellcolor{cyan!30}0.300 & \cellcolor{cyan!25}0.249 & \cellcolor{cyan!20}0.170 & \cellcolor{cyan!20}0.298 & \cellcolor{cyan!25}0.722 & \cellcolor{cyan!15}0.524 & \cellcolor{cyan!20}0.246 & \cellcolor{cyan!15}0.231 \\
$z$ as context input only ($n{=}3$; long $n{=}1$) & \cellcolor{cyan!35}0.307 & \cellcolor{cyan!35}0.301 & \cellcolor{cyan!30}0.253 & \cellcolor{cyan!15}0.174 & \cellcolor{cyan!30}0.300 & \cellcolor{cyan!15}0.705 & \cellcolor{cyan!25}0.550 & \cellcolor{cyan!25}0.250 & \cellcolor{cyan!20}0.228 \\
$z$ as prediction target only ($n{=}3$; long $n{=}1$) & \cellcolor{cyan!30}0.306 & \cellcolor{cyan!30}0.300 & \cellcolor{cyan!30}0.253 & \cellcolor{cyan!15}0.172 & \cellcolor{cyan!20}0.298 & \cellcolor{cyan!20}0.710 & \cellcolor{cyan!20}0.539 & \cellcolor{cyan!30}0.252 & \cellcolor{cyan!25}0.238 \\
\midrule
\multicolumn{10}{l}{\textit{Predictor backbone (random init.)}} \\
ConvLSTM, matched size ($n{=}3$) & \cellcolor{cyan!25}0.303 & \cellcolor{cyan!20}0.297 & \cellcolor{cyan!15}0.236 & \cellcolor{cyan!35}0.134 & \cellcolor{cyan!15}0.296 & \cellcolor{cyan!25}0.717 & \cellcolor{cyan!35}0.592 & \cellcolor{cyan!15}0.239 & \cellcolor{cyan!10}0.197 \\
Per-token head, no mixing ($n{=}3$) & \cellcolor{cyan!10}0.268 & \cellcolor{cyan!10}0.265 & \cellcolor{cyan!10}0.167 & \cellcolor{cyan!20}0.160 & \cellcolor{cyan!10}0.270 & \cellcolor{cyan!10}0.697 & \cellcolor{cyan!10}0.343 & \cellcolor{cyan!10}0.192 & \cellcolor{cyan!10}0.156 \\
\midrule
\multicolumn{10}{l}{\textit{Rollout training (random init.)}} \\
Teacher forcing only & \cellcolor{cyan!15}0.300 & \cellcolor{cyan!15}0.293 & \cellcolor{cyan!15}0.237 & \cellcolor{cyan!25}0.167 & \cellcolor{cyan!20}0.298 & \cellcolor{cyan!10}0.683 & \cellcolor{cyan!30}0.556 & \cellcolor{cyan!15}0.234 & \cellcolor{cyan!15}0.217 \\
Fixed 2 passes & \cellcolor{cyan!25}0.303 & \cellcolor{cyan!20}0.297 & \cellcolor{cyan!20}0.247 & \cellcolor{cyan!10}0.189 & \cellcolor{cyan!15}0.297 & \cellcolor{cyan!20}0.712 & \cellcolor{cyan!20}0.532 & \cellcolor{cyan!15}0.238 & \cellcolor{cyan!15}0.217 \\
\midrule
\multicolumn{10}{l}{\textit{Decoding (random init.)}} \\
Pixel round trip, every step & \cellcolor{cyan!10}0.277 & \cellcolor{cyan!15}0.278 & \cellcolor{cyan!10}0.214 & \cellcolor{cyan!10}0.198 & \cellcolor{cyan!15}0.284 & \cellcolor{cyan!10}0.646 & \cellcolor{cyan!15}0.444 & \cellcolor{cyan!10}0.153 & \cellcolor{cyan!10}0.104 \\
\bottomrule
\end{tabular}}
\end{table*}

\paragraph{EO-WM barely uses the weather it is given.}
We apply the same two controls to EO-WM with its released weights and inference script (Tab. \ref{tab:weather_models}); our run reproduces its reported scores (ENS 0.253 and 0.303, DHR 0.650, PDC 0.285, against 0.254, 0.302, 0.652 and 0.294). EO-WM conditions on daily E-OBS weather anomalies (its five variables minus the tile's monthly climatology), cumulative-stress features and the monthly climatology itself. Withholding its weather, as in its own training-time weather dropout, zeroes the anomalies and the cumulative stress but keeps the climatology, which is the same for both years of a Seasonal pair (same tile and months) and so cannot carry their difference; for EO-JEPA, the whole action is replaced by its learned null embedding, while its date token is kept. This lowers EO-WM's PDC by 0.004 and DHR by 0.014, and giving every window all weather inputs of another window, climatology included, lowers PDC by 0.020; Extreme Summer ENS changes by less than 0.002 under both. The same changes cost EO-JEPA 0.155 and 0.331 in PDC. The response of EO-WM to the two years of a pair is thus carried almost entirely by its context frames, while EO-JEPA's depends on the weather action.

\begin{figure*}[t]
\centering
\includegraphics[width=0.75\linewidth]{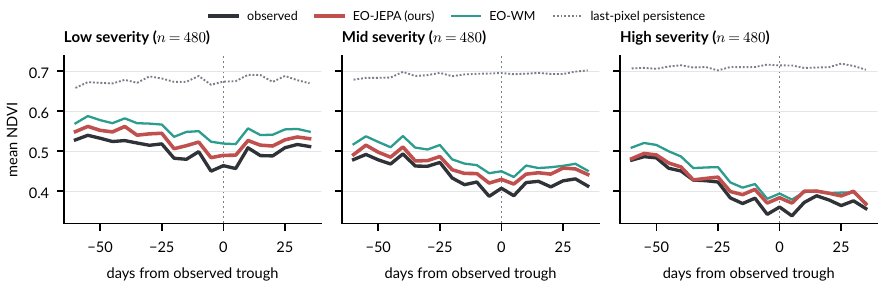}
\caption{\textbf{Vegetation through summer extremes.} Mean NDVI of the cloud-free pixels on the EO-WM Extreme Summer benchmark, with each window aligned on the trough frame of the benchmark, for its three severity tiers (480 windows each; lead times shown where at least 40\% of the windows contribute). EO-JEPA: seed 0, aligned decoder; EO-WM: released weights, ensemble mean of 5 members, run by us (its per-window NDVI errors match the reference output).}
\label{fig:extreme_traj}
\end{figure*}

\paragraph{Extreme Summer trajectories.}
Fig. \ref{fig:extreme_traj} shows the vegetation response, with each window aligned at its observed trough. Across all three severity tiers, EO-JEPA follows the observed decline and recovery more closely than EO-WM, achieving mean absolute gaps of 0.025, 0.022 and 0.016 NDVI, respectively, compared with 0.051, 0.044 and 0.031 for EO-WM. It is closer to observations at 58 of the 60 plotted lead times, while persistence remains at the pre-event level. In the high-severity tier, EO-WM reproduces the decline magnitude slightly better, with an error of 0.150 versus 0.126 for EO-JEPA, against an observed value of 0.148. This is consistent with its lower drop-amplitude error in Tab.~\ref{tab:combined} of the main paper.

\paragraph{The land-surface prior is not what responds.}
Withholding the Tessera prior changes ENS by at most 0.3 points and leaves PDC and DHR unchanged (0.584 and 0.768), so the year-to-year response does not come from a memorized description of the land cover. Without the weather, however, the prior does help: withholding both inputs lowers PDC further, to 0.383. Elevation and date, even withheld together, change ENS by at most 0.002 and PDC by 0.010 (Sec. \ref{supp:design}).

\paragraph{Drop-amplitude error under the controls.}
Two controls lower the drop-amplitude error (DAE) while degrading the forecast. DAE compares the observed drop with the baseline minus the lowest predicted cube-mean NDVI over the forecast, regardless of when or where it occurs, and EO-JEPA underestimates the depth of the drop (at the observed trough, its mean NDVI is 0.49, 0.43 and 0.38 in the three severity tiers, against 0.46, 0.41 and 0.36 observed). Any change that lowers the predicted minimum therefore reduces DAE. With the weather of another location, which on Extreme Summer is the weather of the same 2018 drought at another place and time, DAE falls from 0.232 to 0.223 while NDVI MAE rises by 0.009, trough-NDVI MAE by 0.008 and ENS falls from 0.271 to 0.259 (paired over the 1{,}440 windows; the 95\% bootstrap intervals of the three per-window differences exclude zero). Withholding Tessera likewise lowers DAE (0.217) while raising NDVI and trough-NDVI MAE (by 0.003 and 0.004) and lowering ENS on every benchmark; its DHR gain (0.768 vs.\ 0.760) is within the variation across training seeds ($\pm 0.008$), and PDC is unchanged. We keep the prior because it improves accuracy consistently, if slightly, and because it carries the response when the weather is unavailable.

\paragraph{Direction of year-to-year change.}
Fig.~\ref{fig:change} of the main paper shows the change between the two years of a Seasonal pair map by map. Over all 422 pairs and lead times of 55 to 100 days, EO-JEPA predicts the sign of the observed change (where $|\Delta\mathrm{NDVI}|\geq 0.1$) on 75.4\% of pixels; withholding the weather lowers this to 71.2\% and the weather of another location to 61.2\%, while EO-WM stays at 68.6\% under all three inputs. On the pairs whose years differ most (top quartile of observed NDVI divergence, pooled over pixels and lead times, so not comparable with the per-pair 74\% of Fig.~\ref{fig:change}), EO-JEPA's direction accuracy changes little when the weather is withheld (79.8\% vs.\ 78.6\%), as the two years already differ in their context frames.

\begin{figure}[t]
\centering
\includegraphics[width=\linewidth]{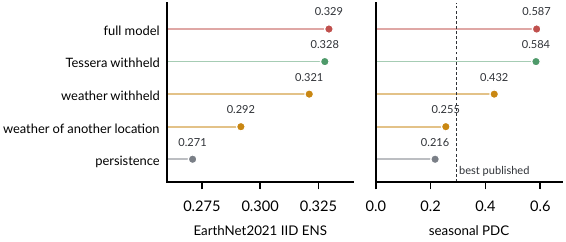}
\caption{\textbf{EarthNet2021 ENS barely registers the weather; the weather response does.} The same trained model with inputs changed at inference (Tab. \ref{tab:controls}). Withholding the weather costs 0.8 ENS points on IID but a quarter of the seasonal pair-difference correlation (PDC); the dashed line marks the best published PDC.}
\label{fig:response}
\end{figure}

\paragraph{Implication for evaluation.}
Withholding the weather changes the EarthNet2021 IID score by less than one point while it removes a quarter of the matched-pair response. Standard accuracy scores therefore cannot tell whether a forecaster follows the weather; the response metrics can, and they are where latent world modeling differs most from prior models.
\begin{table*}[t]
\centering
\caption{\textbf{Predictor initialization with aligned decoders.} Mean and standard deviation over three training seeds (a superscript gives the number of seeds where fewer are available); seeds 1 and 2 are decoded with the decoder aligned to seed 0 of the same initialization.}
\label{tab:supp_init}
\setlength{\tabcolsep}{4pt}
\resizebox{\linewidth}{!}{%
\begin{tabular}{l|cc|cc|ccc|cc}
\toprule
\multirow{2}{*}{Variant} & \multicolumn{2}{c|}{EarthNet2021} & \multicolumn{2}{c|}{Extreme Summer} & \multicolumn{3}{c|}{Seasonal pairs} & \multicolumn{2}{c}{Long horizon} \\
 & IID$\uparrow$ & OOD$\uparrow$ & ENS$\uparrow$ & DAE$\downarrow$ & ENS$\uparrow$ & DHR$\uparrow$ & PDC$\uparrow$ & 200\,d$\uparrow$ & 700\,d$\uparrow$ \\
\midrule
V-JEPA~2 (EO-JEPA) & \cellcolor{cyan!30}$0.330_{\pm 0.001}$ & \cellcolor{cyan!30}$0.321_{\pm 0.001}$ & \cellcolor{cyan!35}$0.272_{\pm 0.001}$ & \cellcolor{cyan!35}$0.228_{\pm 0.010}$ & \cellcolor{cyan!30}$0.318_{\pm 0.001}$ & \cellcolor{cyan!35}$0.756_{\pm 0.008}$ & \cellcolor{cyan!35}$0.594_{\pm 0.014}$ & \cellcolor{cyan!15}$0.282_{\pm 0.002}$ & \cellcolor{cyan!15}$0.262_{\pm 0.001}$ \\
Random & \cellcolor{cyan!35}$0.331_{\pm 0.000}$ & \cellcolor{cyan!35}$0.322_{\pm 0.000}$ & \cellcolor{cyan!35}$0.272_{\pm 0.000}$ & \cellcolor{cyan!15}$0.245_{\pm 0.001}$ & \cellcolor{cyan!35}$0.319_{\pm 0.000}$ & \cellcolor{cyan!15}$0.739_{\pm 0.002}$ & \cellcolor{cyan!15}$0.563_{\pm 0.027}$ & \cellcolor{cyan!35}$0.286_{\pm 0.001}$ & \cellcolor{cyan!35}$0.270_{\pm 0.003}$ \\
\bottomrule
\end{tabular}}
\end{table*}
\begin{table*}[t]
\centering
\caption{\textbf{Results by land cover.} Each test cube is assigned the ESA WorldCover 2021 class that covers most of its pixels (median share 68\%); classes with at least 50 cubes or pairs are shown. Long-horizon columns: official EarthNetScore within the class (EO-JEPA: mean of three seeds). Seasonal pairs: PDC (Spearman over the class's pairs) and DHR. Extreme Summer is not split, as 1{,}374 of its 1{,}440 windows are cropland. Gray: last-pixel persistence.}
\label{tab:supp_landcover}
\scriptsize
\setlength{\tabcolsep}{2pt}
\resizebox{\textwidth}{!}{%
\begin{tabular}{l|ccccc|ccccc|cccccc}
\toprule
 & \multicolumn{5}{c|}{EarthNet2021, 200\,d ENS$\uparrow$} & \multicolumn{5}{c|}{EarthNet2021, 700\,d ENS$\uparrow$} & \multicolumn{6}{c}{Seasonal pairs} \\
 & & & & & & & & & & & & \multicolumn{3}{c}{PDC$\uparrow$} & \multicolumn{2}{c}{DHR$\uparrow$} \\
Class & $n$ & EO-JEPA & SGConvLSTM & ConvLSTM & Pers. & $n$ & EO-JEPA & SGConvLSTM & ConvLSTM & Pers. & $n$ & EO-JEPA & EO-WM & Pers. & EO-JEPA & EO-WM \\
\midrule
All & 4{,}000 & \cellcolor{cyan!20}0.282 & \cellcolor{cyan!20}0.274 & \cellcolor{cyan!20}0.214 & \textcolor{gray}{0.183} & 4{,}000 & \cellcolor{cyan!25}0.262 & \cellcolor{cyan!25}0.222 & \cellcolor{cyan!25}0.219 & \textcolor{gray}{0.210} & 422 & \cellcolor{cyan!25}0.587 & \cellcolor{cyan!30}0.285 & \textcolor{gray}{0.216} & \cellcolor{cyan!20}0.760 & \cellcolor{cyan!20}0.650 \\
\midrule
Tree cover & 1{,}521 & \cellcolor{cyan!35}0.311 & \cellcolor{cyan!35}0.304 & \cellcolor{cyan!35}0.219 & \textcolor{gray}{0.254} & 1{,}475 & \cellcolor{cyan!35}0.291 & \cellcolor{cyan!25}0.222 & \cellcolor{cyan!30}0.224 & \textcolor{gray}{0.254} & 175 & \cellcolor{cyan!20}0.503 & \cellcolor{cyan!25}0.231 & \textcolor{gray}{0.251} & \cellcolor{cyan!10}0.699 & \cellcolor{cyan!10}0.625 \\
Cropland & 2{,}335 & \cellcolor{cyan!10}0.262 & \cellcolor{cyan!10}0.255 & \cellcolor{cyan!10}0.211 & \textcolor{gray}{0.104} & 1{,}214 & \cellcolor{cyan!10}0.207 & \cellcolor{cyan!10}0.183 & \cellcolor{cyan!10}0.182 & \textcolor{gray}{0.106} & 152 & \cellcolor{cyan!10}0.398 & \cellcolor{cyan!35}0.317 & \textcolor{gray}{0.310} & \cellcolor{cyan!15}0.752 & \cellcolor{cyan!15}0.645 \\
Grassland & 142 & \cellcolor{cyan!15}0.268 & \cellcolor{cyan!15}0.262 & \cellcolor{cyan!25}0.216 & \textcolor{gray}{0.198} & 1{,}278 & \cellcolor{cyan!30}0.269 & \cellcolor{cyan!35}0.250 & \cellcolor{cyan!35}0.241 & \textcolor{gray}{0.218} & 82 & \cellcolor{cyan!35}0.727 & \cellcolor{cyan!10}0.060 & \textcolor{gray}{0.084} & \cellcolor{cyan!35}0.883 & \cellcolor{cyan!35}0.696 \\
\bottomrule
\end{tabular}}
\end{table*}

\section{Rolling out in latent space}
\label{supp:rollout}
Rolling out in latent space, with decoding deferred to the output, is what keeps long rollouts stable and the weather response faithful. The pixel round trip tests this directly on the trained model: after every predicted step, the predicted EO-VAE tokens are decoded to reflectance and re-encoded by the frozen encoders before being fed back, so the rollout passes through pixel space while the model, its weights and its inputs stay the same (Tab.~\ref{tab:controls}, bottom). Routing every step through pixels lowers ENS on every benchmark, by 1.9 points on IID, 1.4 on Extreme Summer and 1.6 on the Seasonal pairs, and PDC from 0.587 to 0.511. The cost compounds with the horizon: 2.8 points at 200 days (0.282 to 0.254) and 6.4 at 700 days (0.263 to 0.199), where it falls below both persistence baselines. Fig.~\ref{fig:lead} of the main paper shows where the two rollouts separate. Re-encoding only between the 100-day chunks (once at 200 days, six times at 700 days) is harmless (0.282 at 200 days, 0.269 at 700 days). It is therefore the per-step pixel feedback that breaks the rollout, not decoding as such: once errors in reflectance (clouds, texture, the decoder's own reconstruction error) are fed back at every step, they accumulate, whereas the latent rollout carries only what the encoders keep. The randomly initialized model loses more under the per-step round trip (Tab.~\ref{tab:ablation}, shared decoder: 3.0 points on IID, PDC 0.444), so the effect does not depend on the pretrained weights.

In the NDVI trajectories of Fig.~\ref{fig:supp_lead},} a ConvLSTM predictor trained in our latent space achieves comparable average errors to EO-JEPA, at 0.102 and 0.115 versus 0.106 and 0.117, but develops an upward bias late in the second year, averaging 0.079 over the final 100 days, compared with an underestimation of 0.030 for EO-JEPA. Across architectures, the same pattern holds. A ConvLSTM predictor of matched size trained in our latent space (Tab.~\ref{tab:ablation}) reaches a PDC of 0.592 and a DHR of 0.717, whereas the pixel-space recurrent models evaluated by EO-WM reach $-0.150$ (PredRNN) and 0.123 (PredRNN~v2) (Tab.~\ref{tab:combined}). Although these models differ in more than the space they forecast in, the gap of 0.47--0.74 in PDC between recurrent models in latent and in pixel space is an order of magnitude larger than the difference between our transformer and ConvLSTM predictors in the same latent space (0.035).

\section{Results by land cover}
\label{supp:landcover}
To see where the model works best, we assign each test cube the ESA WorldCover 2021 class~\cite{zanaga2022esa} that covers most of its pixels and recompute the main metrics within the three classes that are sufficiently represented: tree cover, cropland and grassland (Tab.~\ref{tab:supp_landcover}). EO-JEPA is the best learned model in every class on both long-horizon tracks and on the Seasonal pairs. Its margin is largest where vegetation follows the weather most directly: on grassland, its PDC is 0.727 against 0.060 for EO-WM, and in tree cover its 700-day ENS exceeds the best published model by 6.7 points (0.291 vs.\ 0.224). Cropland is the hardest class for all models (700-day ENS 0.207) and the one where the weather response is closest to persistence (PDC 0.398, against 0.317 for EO-WM and 0.310 for persistence): sowing, harvest and crop rotation change the surface independently of the weather, and no input of the model describes them. Part of the overall PDC reflects differences between classes; within tree cover and cropland it is lower than over all pairs.

\section{Component ablations}
\label{supp:ablations}

Tab.~\ref{tab:ablation} isolates the components of the model. All rows decode with the same decoder, so the differences come from the predictor; rows marked with $n$ average that many training seeds.

\paragraph{Semantic stream.}
Removing the DOFA stream $z$, so that the model forecasts only the decodable EO-VAE tokens, leaves both EarthNet2021 scores unchanged ($0.307/0.300$ vs.\ $0.307/0.301$) but lowers Extreme Summer ENS by 0.4 points and PDC by 0.033, and each of the three seeds without $z$ is below each of the three seeds with it on both metrics. Either role of $z$ alone recovers most of this: keeping it only as a prediction target, or only as a context input, restores the Extreme Summer score (0.253 for both, three seeds each), and PDC is highest when $z$ is both input and target (0.557, against 0.550 as input only, 0.539 as target only and 0.524 without $z$). On the 700-day track, single runs favor the target-only variant (0.238 vs.\ 0.228 as input only), a difference we do not consider resolved. The semantic stream thus helps under extremes and for the weather response, and the evidence does not single out its use as a prediction target.

\paragraph{Rollout training.}
Training with teacher forcing alone lowers ENS on every track: by 0.7 points on IID and 1.6 on Extreme Summer against the three-seed mean, and by 1.5 points on both the 200- and 700-day tracks against the same seed (0.234 and 0.217 vs.\ 0.249 and 0.232; against the three-seed means of Tab.~\ref{tab:ablation}, 0.249 and 0.225, the drops are 1.5 and 0.8 points). A fixed two-pass schedule recovers part of the 100-day scores but not the long-horizon ones (0.238 and 0.217). Exposing the predictor to its own predictions, with a curriculum to deeper rollouts, is what makes chunked rollouts over hundreds of days viable.

\paragraph{Predictor.}
A per-token head without spatial or temporal mixing loses 3.9 ENS points on IID, 8.6 on Extreme Summer and 6.9 on the 700-day track, and its PDC falls to 0.343. A ConvLSTM of matched size is close on EarthNet2021 (0.303 vs.\ 0.307) but 1.7 points lower on Extreme Summer and 2.8 lower on the 700-day track; it responds slightly more strongly to the weather (PDC 0.592 vs.\ 0.557, DAE 0.134 vs.\ 0.170). The transformer predictor is thus the more accurate forecaster under extremes and over long horizons, while the weather response is a property of the latent formulation shared by both predictors.

\paragraph{Initialization.}
Initializing the predictor from V-JEPA~2 does not improve on random initialization with the shared decoder (Extreme Summer ENS 0.245 vs.\ 0.253, PDC 0.548 vs.\ 0.557, three seeds each). With aligned decoders (Tab.~\ref{tab:supp_init}), the V-JEPA-initialized model has the stronger weather response (PDC $0.594 \pm 0.014$ vs.\ $0.563 \pm 0.027$, DHR 0.756 vs.\ 0.739, drop-amplitude error 0.228 vs.\ 0.245), and the randomly initialized one has equal or slightly higher ENS on every track (IID 0.331 vs.\ 0.330, Extreme Summer 0.272 for both; 0.286 vs.\ 0.282 at 200 days and 0.270 vs.\ 0.262 at 700 days, three seeds each). The gains come from learning latent next-state dynamics rather than from the pretrained weights.



\end{document}